\documentclass[10pt,journal,compsoc]{IEEEtran}
\usepackage{graphicx}
\usepackage{url}
\usepackage[table]{xcolor}
\usepackage{multirow}
\usepackage{tikz}
\usepackage{makecell}
\usepackage{booktabs}
\usepackage{subcaption}
\usepackage[backgroundcolor=lightgray]{todonotes}
\usepackage{soul}

\definecolor{Gray}{gray}{0.2}
\definecolor{LimeGreen}{rgb}{0.00, 0.50, 0.00}
\definecolor{lightgray}{gray}{0.7}

\newcommand{\argmax}{\mathop{\mathrm{arg\,max}}}
\newcommand{\miqueldel}[1]{\textcolor{red}{}}
\newcommand{\miqueladd}[1]{#1}
\newcommand{\revadd}[1]{#1}

\ifCLASSOPTIONcompsoc
  \usepackage[nocompress]{cite}
\else
  \usepackage{cite}
\fi

\begin{document}

\title{Non-fluent synthetic target-language data improve neural machine translation}

\author{Víctor M. Sánchez-Cartagena,~
        Miquel Esplà-Gomis,~
        Juan Antonio Pérez-Ortiz,~
        Felipe Sánchez-Martínez~
\IEEEcompsocitemizethanks{
\IEEEcompsocthanksitem V.M.~Sánchez-Cartagena, M.~Esplà-Gomis, J.A.~Pérez-Ortiz and F.~Sánchez-Martínez are with the Dep. de Llenguatges i Sistemes Informàtics, Universitat d’Alacant, Spain.\protect\\
E-mail: \{vmsanchez,mespla,japerez,fsanchez\}@dlsi.ua.es}

\thanks{\textup{V.M. Sánchez-Cartagena, M. Esplà-Gomis, J.A. Pérez-Ortiz and F.~Sánchez-Martínez
(2023), ``Non-fluent synthetic target-language data improve neural machine translation,'' in 
\emph{IEEE Transactions on Pattern Analysis and Machine Intelligence}, Vol. 46, No. 2, pp. 837--850, doi: 10.1109/TPAMI.2023.3333949.\newline Accepted version.
Published online: November 2023.\newline \texttt{https://doi.org/10.1109/TPAMI.2023.3333949}}}
}

\markboth{IEEE TRANSACTIONS ON PATTERN ANALYSIS AND MACHINE INTELLIGENCE}%
{Sánchez-Cartagena \MakeLowercase{\textit{et al.}}: Non-fluent synthetic target-language data improve neural machine translation}

\IEEEtitleabstractindextext{%
\begin{abstract}
When the amount of parallel sentences available to train a neural machine translation is scarce, a common practice is to generate new synthetic training samples from them. A number of approaches have been proposed to produce synthetic parallel sentences that are similar to those in the parallel data available. These approaches work under the assumption that non-fluent target-side synthetic training samples can be harmful and may deteriorate translation performance. Even so, in this paper we demonstrate that synthetic training samples with non-fluent target sentences can improve translation performance if they are used in a multilingual machine translation framework as if they were sentences in another language.  We conducted experiments on ten low-resource and four high-resource translation tasks and found out that this simple approach consistently improves translation performance as compared to state-of-the-art methods for generating synthetic training samples similar to those found in corpora. Furthermore, this improvement is independent of the size of the original training corpus, the resulting systems are much more robust against domain shift and produce less hallucinations.
\end{abstract}

}

\maketitle

\IEEEdisplaynontitleabstractindextext
\IEEEpeerreviewmaketitle

\IEEEraisesectionheading{
\section{Introduction}\label{sec:introduction}
}

\IEEEPARstart{M}{achine} translation (MT) ---the application of computers to the task of translating a text in one natural language into another without human intervention---  is one of the key technologies enabling communication in our globalized world. Millions of users rely on MT on a daily basis for either \emph{assimilation}, the use of the raw MT output to get an idea of the meaning of texts in languages they do not understand, or for \emph{dissemination}, the use of the MT output to create a draft translation which is then manually corrected and published.

The uptake of MT technology has gradually increased over the last ten years \cite{hassan2018achieving}, mainly motivated by the recent advances in the state-of-the-art approach to MT, namely \emph{neural machine translation} (NMT). NMT models are data intensive and require large amounts of \emph{parallel corpora} in the form of \miqueladd{several hundreds of thousands or} millions of human-translated sentence pairs used for their training. Besides that, monolingual data has also proven to be a valuable resource to train NMT systems. 

Although there exist language pairs, such as English--German or English--Spanish, that have large, freely available parallel corpora, most language pairs may be considered \emph{low-resource} because there is little or no translated text available to train NMT models for them. This problem has been addressed in NMT through different approaches, such as transfer-learning from high-resource language pairs~\cite{kocmi-bojar-2018-trivial}, using linguistic annotations~\cite{sennrich-haddow-2016-linguistic}, 
training multilingual systems~\cite{doi:10.1162} and applying data augmentation strategies~\cite{li-etal-2019-understanding-data,survey2021}, i.e., artificially generating additional parallel sentences.

Data augmentation (DA) is formalized by many authors as a solution to a data distribution mismatch problem~\cite{wang-etal-2018-switchout,wei-etal-2020-uncertainty} in which the empirical data distribution of the sentence pairs in the training corpus differs from the true data distribution. DA is therefore used to build unseen synthetic training samples that are plausible under the true data distribution. 
\miqueladd{When these synthetic training samples are used during training, the resulting augmented data distribution is expected to be closer to the true data distribution.}
The final objective is to obtain models that are able to properly translate sentences \miqueladd{plausible} under the true data distribution, even though they may be unlikely under the empirical data distribution \miqueladd{in the training corpus}. 

\miqueladd{For this reason, most DA approaches}
 pay special attention to avoid\miqueladd{ing} non-fluent target-language sentences in the synthetic training samples~\cite{sennrich-back,wang-etal-2018-switchout}, as their use may lead to worse NMT models with lower translation performance.
In this paper, however, and in contrast to previous works, we describe a DA approach that consists of  building synthetic training samples that are deliberately implausible under the \miqueladd{true} data distribution.  The objective is to strengthen the encoder of the NMT model through DA. Voita et al.~\cite{voita2020analyzing} claim that the influence of encoder representations in the output predictions of an NMT system is higher when the NMT system is trained on large corpora than when it is trained on small corpora.
By producing  \emph{unlikely} synthetic training samples, especially as regards their target side, we aim at exposing the network to new situations where the target-language prefix \miqueladd{does not provide sufficient context} to predict the next token, therefore forcing the decoder to rely more on the encoder representations for its predictions; \miqueladd{this should be viewed as a desirable trait since MT systems should build upon the source sentence to produce accurate translations.} In this way, it is possible to build NMT models for low-resource language pairs that, even though they have been trained on small parallel corpora, are able to  behave as if they had been trained on larger training corpora.

Obviously, synthetic training samples with non-fluent target-language sentences cannot be used as if they were original training samples, since their use would harm the target-language model learned by the NMT system. To avoid this, we propose the use \miqueladd{of} a multi-task learning framework during training. 
This is easily achieved, without changing the model architecture, by \miqueladd{prepending} a task-specific artificial token to the source sentence to constrain the kind of output to be produced~\cite{sennrich2016token,doi:10.1162}, similarly to what is done in one-to-many multilingual NMT~\cite{dong-etal-2015-multi} to specify the target language. 
We term this DA approach as \emph{multi-task learning data augmentation}~(MaTiLDA).




Our framework ---which extends preliminary work\footnote{The \miqueladd{additional} contributions of this paper with respect to the conference paper~\cite{sanchez-cartagena21a} are as follows:
(i) more sophisticated approach for generating the synthetic samples during training;
(ii) additional experiments on both low-resource and high-resource translation tasks;
(iii) more exhaustive comparison to other DA methods;
(iv) improved evaluation of the contribution of the source representations generated by the encoder during inference; and
(v) better grounded analysis of the tendency to hallucinate of the models evaluated.} reported in a conference paper by the same authors \cite{sanchez-cartagena21a}--- does not require elaborate preprocessing steps, training additional systems, or data besides the available training parallel corpora. Experiments with ten low-resource translation tasks show that it systematically outperforms state-of-the art methods aimed at extending the support of the empirical data distribution. Additional experiments on four high-resource translation tasks show that this approach is also able to improve translation performance even in high-resource conditions in which NMT systems are trained on large parallel corpora. Furthermore, we show that this new approach and the standard DA approach, back-translation~\cite{sennrich-back}, complement each other and allow further performance improvements when they are used together.

In addition to these experiments, we perform an analysis of the relevance of the encoder and decoder representations in the NMT system output, which shows that, thanks to the 
added 
synthetic training samples, MaTiLDA increases the contribution of the source representations generated by the encoder to the decisions made by the NMT decoder during inference. Moreover, systems trained with MaTiLDA are much more robust against domain shift, and produce less hallucinations~\cite{wang-sennrich-2020-exposure}. 


The remainder of the paper is organized as follows. The next section briefly describes the neural approach to MT. After that, Sec.~\ref{sec:da} describes the DA  strategies we follow and evaluate in our experiments and the modifications introduced to the training process of the NMT system. Sec.~\ref{sec:settings} then describes the experimental settings, whereas Sec.~\ref{sec:results} \miqueladd{reports} and discusses the results obtained on low-resource and high-resource translation tasks. Sec.~\ref{sec:explain} presents an analysis of the changes in the \miqueladd{use of the encoder representations} induced by our DA strategies, \revadd{and an analysis of the tendency to hallucinate of the systems trained with MaTiLDA}.
The paper ends with a review of the most relevant works in the area of DA for NMT in Sec.~\ref{sec:related}, followed by some concluding remarks in Sec.~\ref{sec:conclusions}.

\section{Neural machine translation}\label{sec:nmt}
Given a source sentence, $\mathbf{x} = \langle x_1, ..., x_n \rangle$, and its translation, $\mathbf{y} = \langle y_1, ...,y_m \rangle$, NMT systems factorize the translation probability $p(\mathbf{y}|\mathbf{x})$ as:
\begin{equation}
    p(\mathbf{y}|\mathbf{x})= \prod_{j=1}^m p(y_j|\mathbf{y_{<j}},\mathbf{x}) \mbox{,}
\label{eq:factorisation}
\end{equation}
where $\mathbf{y_{<j}}$ stands for the target prefix produced before predicting the $j$-th token, $y_j$, in the target language. 

Different families of neural networks have been proposed for producing this probability distribution: convolutional~\cite{gehring-etal-2017-convolutional}, recurrent~\cite{bahdanau2015neural} and transformer~\cite{vaswani2017attention},
the latter \miqueladd{being} the current state of the art.
These neural networks are optimized by seeking the model parameters $\theta^*$ that maximize the likelihood of the training data $\mathcal{D}$:

\begin{equation}
    \theta^* = \argmax_{\theta} \sum_{\langle \mathbf{x},\mathbf{y} \rangle \in \mathcal{D}} p(\mathbf{y}|\mathbf{x}; \theta) \mbox{.}
\end{equation}

Stochastic gradient-based optimization methods are commonly applied to find a local maximum of \miqueladd{that} likelihood. These optimization methods involve iteratively updating the model parameters using only a mini-batch of sentences $\mathcal{B}$ 
from the training corpus. The parameters $\theta$ are updated in terms of the gradient of a loss $\mathcal{L}(\mathcal{B},\theta)$ computed over the mini-batch. For instance, parameters in the batched stochastic gradient descent  algorithm~\cite{robbins1951stochastic} are updated as follows, where $i$ is the iteration index, $n$ is the number of target words in the mini-batch $\mathcal{B}$ and $\eta$ is the learning rate:
\begin{equation}
    \theta_{i+1} = \theta_{i} - \eta \frac{1}{n} \nabla \mathcal{L}(\mathcal{B},\theta) \mbox{.}
\end{equation}

NMT systems are usually optimized using a loss based on cross-entropy. The basic formulation of the cross-entropy loss function is \miqueladd{shown} below, although some enhancements such as label smoothing have become increasingly popular~\cite{muller2019does}:

\begin{equation}
    \mathcal{L}(\mathcal{B},\theta) = \sum_{\langle \mathbf{x},\mathbf{y} \rangle \in \mathcal{B}} \log p(\mathbf{y}|\mathbf{x}; \theta) \mbox{.}
\label{eq:loss}
\end{equation}

\section{Data augmentation strategies}\label{sec:da}
\begin{table*}[tb]
\centering
\fontsize{10.2}{12.0}\selectfont
\begin{tabular}{lll}
\textbf{Task} & \textbf{Lang.} & \textbf{Synthetic training sample} \\
\hline \hline
\rowcolor{gray!10}
 & source & Es gibt andere Möglichkeiten , die Pyramide zu durchbrechen . \\
 \rowcolor{gray!10}
                            & & \\
                            \rowcolor{gray!10}
                            & & \multirow{-2}{*}{
                           \begin{tikzpicture}
                               \draw (0.5,0) -- (1.6,-1);
                               \draw (1.2,0) -- (0.9,-1);
                               \draw (2.0,0) -- (2.1,-1);
                               \draw (3.8,0) -- (2.9,-1);
                               \draw (5.5,0) -- (5.7,-1);
                               \draw (6.5,0) -- (6.7,-1);
                               \draw (7.7,0) -- (3.1,-1);
                               \draw (8.7,0) -- (4.3,-1);
                               \draw (10.4,0) -- (7.1,-1);
                            \end{tikzpicture} }\\
\rowcolor{gray!10}
\multirow{-4}{*}{\makecell[l]{original\\ training\\ sample}}  & target & There 's other ways of breaking the pyramid . \\ \midrule
swap & target & There \textcolor{red}{.} other ways of breaking \textcolor{blue}{pyramid} \textcolor{red}{'s} \textcolor{blue}{the} \\ \hline
unk & target & There 's other \textcolor{red}{UNK} of \textcolor{red}{UNK UNK UNK } . \\ \hline
source & target & \textcolor{red}{Es gibt andere Möglichkeiten , die Pyramide zu durchbrechen .} \\ \hline
reverse & target & \textcolor{red}{. pyramid the breaking of ways other 's There} \\ \hline
mono & target & \textcolor{red}{'s There} other ways \textcolor{red}{the pyramid of breaking} . \\ \hline
\multirow{2}{*}{replace} & source & Es gibt \textcolor{red}{aufzurüsten} \textcolor{blue}{kalt} , \textcolor{violet}{Schach} \textcolor{orange}{Spezialwissen} zu durchbrechen . \\ 
                      &  target & There 's \textcolor{red}{arming} \textcolor{blue}{cold} of breaking \textcolor{violet}{chess} \textcolor{orange}{specialties} . \\ 

\hline
\end{tabular}
\caption{A German--English, word-aligned training sample (first row) and the result of applying the transformations described in Sec.~\ref{sec:da} using $\alpha=0.5$ for those transformations controlled by this hyperparameter. Words modified by each transformation are coloured; for \emph{swap} and \emph{replace}, a different colour identifies the pair of words that are either swapped or replaced together, respectively.}
\label{tab:auxtasks}
\end{table*}

The DA strategies we follow can be formalised as transformations that are applied to the original training samples to produce synthetic training samples. These synthetic samples are expected to force the NMT system to rely more on the source-language representations generated by the encoder during translation. Most of the transformations described next produce synthetic samples with non-fluent target-language sentences. Some transformations are controlled by a hyperparameter $\alpha$ that determines the \miqueladd{fraction} of target words affected by the transformation. In what follows, $m$ denotes the amount of words in the original target sentence. Table~\ref{tab:auxtasks} provides an example of the effect of the different transformations on a single sentence pair.

\makeatletter
\newenvironment{mydescription}%
               {\list{}{\itemsep=1.8pt \leftmargin=7pt
                        \labelwidth\z@ \itemindent-\leftmargin
                        \let\makelabel\descriptionlabel}}%
               {\endlist}
\makeatother

\begin{mydescription}
\item[swap:] Pairs of random target words are swapped until only $(1-\alpha)\cdot m $ words remain in their original position. This transformation~\cite{artetxe2017unsupervised}  tries to force the system to trust less the target prefix when generating a new token.

\item[unk:] $\alpha \cdot m$ random target words are replaced by a special UNK token~\cite{xie2017data} when they are fed to the neural network as previous context ($\mathbf{y_{<j}}$; see Eq.~\ref{eq:factorisation}) for the prediction of the next target token $y_j$. Note that, when computing the training loss, the original token, rather than the UNK token, is used as the expected output. This strategy is similar to the word dropout used when preventing posterior collapse in variational autoencoders~\cite{bowman-etal-2016-generating} and makes the target prefix less informative for the prediction of the next target token.  


\item[source:] The target sentence becomes a copy of the source sentence. Thus, the most efficient way of \miqueladd{predicting} the right output is checking the encoder representation to copy from the source. Although 
such training instances have been identified
as harmful~\cite{pmlr-v80-ott18a,khayrallah-koehn-2018-impact}, we empirically found the opposite (see Sec.~\ref{sec:results}) thanks to the multi-task learning framework defined later in this section.



\item[reverse:] The order of the words in the target sentence is reversed. Voita et al.~\cite{voita2020analyzing}
suggest that the influence of the encoder decreases along the target sentence; therefore, by reversing the order we expect the system to learn to use more information from the encoder when generating tokens that usually appear near the end of the sentence. 

\item[mono:] Target words are reordered to make the alignment between source and target words monotonous by using one-to-many word alignments as in the compression of parallel corpora~\cite{sanchez-martinez12a}. By making the alignment between source and target words monotonous, the target sentences become less fluent, so we expect the system to pay more attention to the encoder.

\item[replace:] $\alpha \cdot m$ source--target aligned \miqueladd{pairs of} words are selected at random and replaced by random entries in a bilingual lexicon obtained from the training corpus; to this end, one-to-one word alignments are used.\footnote{If the number of aligned words is below $\alpha \cdot m$, all available alignments are used.} This transformation is likely to introduce words that are difficult to produce by relying only on the target language prefix, thus forcing the system to pay attention to the source words. Fadaee et al.~\cite{Fadaee_2017} followed a similar approach; however, they constrained the replacements to produce only fluent target sentences.
\end{mydescription}

Using the original training samples together with the synthetic ones, without distinction between them, would degrade the overall translation performance. On the one hand, the system would not be able to learn the kind of output to be produced (e.g. well-formed, fluent target sentences); on the other hand, the system could learn spurious correlations from the synthetic training samples. In order to minimize the negative impact of having non-fluent target sentences in the synthetic training samples while keeping their ability to force reliance on the encoder, we applied the multi-task learning strategy described next.

We organize the original training data in mini-batches as if a vanilla NMT system were trained. Then, for each original sentence pair $\langle \mathbf{x},\mathbf{y} \rangle$ in a mini-batch, we apply one or more of the transformations aforementioned, depending on the experimental setting. For each transformation $i$, a new synthetic sentence pair $\langle \mathbf{\hat{x}^i},\mathbf{\hat{y}^i} \rangle$ is generated, and a new term is added to the loss function to account for its  cross-entropy.\footnote{Note that in all transformations but \emph{replace}, $\mathbf{\hat{x}^i}=\mathbf{x}$.} As mini-batches are created at the beginning of each epoch, the result of applying the transformations that involve random decisions is different for each epoch, thus preventing the system from learning spurious correlations from synthetic data.\footnote{This differs from previous work~\cite{sanchez-cartagena21a} in which transformations were applied to the original training samples during the pre-processing of the corpus, and therefore before training.} In addition, we add a token to each source sentence to indicate whether it is part of an original training sample or of synthetic one, and, in the latter case, which transformation was used for its generation. The latter resembles what is done in multilingual NMT~\cite{sennrich2016token,doi:10.1162} to indicate the language of the output to be produced.

Hence, within our MaTiLDA framework, and with $r$ transformations, the cross-entropy loss function described in Eq.~(\ref{eq:loss}) turns into the following expression:

\begin{equation}\label{eq:loss2tasks}
\mathcal{L}(\mathcal{B},\theta) =  \sum_{\langle \mathbf{x},\mathbf{y} \rangle \in \mathcal{B}} \frac{1}{r+1} \sum_{i=0}^r \log p(\mathbf{\hat{y}^i}|\langle t_i, \hat{x}_1^i, ..., \hat{x}_{\hat{n}_i}^i \rangle; \theta)
\end{equation}
where $\langle \mathbf{\hat{x}^0},\mathbf{\hat{y}^0} \rangle$ corresponds to the original training sample
$\langle \mathbf{x},\mathbf{y} \rangle$, $\hat{n_i}$ stands for the length of  $\mathbf{\hat{x}^i}$, and $t_{i}$ denotes the artificial token introduced to indicate whether the training sample is the original one ($i=0$) or the synthetic one corresponding to the $i$-th transformation ($i \ge 1$). 



\section{Experimental settings}
\label{sec:settings}
\miqueladd{We have conducted experiments with five language pairs in a low-resource scenario: English (en) into German (de), Hebrew (he) and Vietnamese (vi), and German into Upper Sorbian (hsb) and Romansh (rm). For this purpose, a collection of corpora commonly considered for evaluating DA techniques in low-resource conditions were used (see Sec.~\ref{se:datasets}).} In addition, we have conducted experiments  \miqueladd{with} two high-resource language pairs: English into Romanian (ro) and English into German. \miqueladd{Both translation directions were evaluated for each language pair, which makes a total of ten low-resource translation tasks and four high-resource translation tasks.} 

We evaluated the effect of using each of the MaTiLDA transformations, as well as the combination of the best performing ones. 
\miqueladd{We explored the combination of MaTiLDA and back-translation, }
and we also evaluated four strong DA methods that aim at extending the support of the empirical data distribution: SwitchOut \cite{wang-etal-2018-switchout}, RAML \cite{norouzi2016reward}, the combination of both (SwOut+RAML) and SeqMix \cite{guo-etal-2020-sequence}. SwitchOut and RAML replace some words by random samples from the vocabulary: SwitchOut works on the source language, and RAML on the target language. 
SeqMix creates synthetic training samples by randomly combining parts of two sentences. 

\subsection{Datasets} \label{se:datasets}
For the experiments in low-resource conditions on five different language pairs we used the following corpora.
For English--German and English--Hebrew, we followed Guo et al.~\cite{gao2019soft} and Gao et al.~\cite{guo-etal-2020-sequence}, and used the training data (speeches of TED and TEDx talks) of the IWSLT 2014 text translation track~\cite{iwslt2014};\footnote{\url{https://sites.google.com/site/iwsltevaluation2014/data-provided}} for development and testing we used the \emph{tst2013} and \emph{tst2014} datasets, respectively. For
English--Vietnamese we used, like Wang et al.~\cite{wang-etal-2018-switchout}, the training data (also TED talks) of the IWSLT 2015 text translation track~\cite{iwslt2015};\footnote{\url{https://wit3.fbk.eu/2015-01}} datasets \emph{tst2012} and \emph{tst2013} were used, respectively, for development and testing. For German--Upper Sorbian we used the corpora released as part of the WMT 2021 task on very low resource supervised machine translation;\footnote{\url{https://www.statmt.org/wmt21/unsup_and_very_low_res.html}} for training we used all the corpora allowed for the task; for development and testing we used the \emph{devel} and \emph{devel\_test} sets, respectively, provided by the organizers of the task. For German--Romansh we used the law corpus released by Müller et al.~\cite{muller-etal-2020-domain} as well as the same split into training, development and testing they used.


For the combination of MaTiLDA and 
back-translation \cite{sennrich-back}, we used additional English and German monolingual corpora. In particular, for English we used all the available monolingual English sentences in the IWSLT 2018 shared task on low-resource MT of TED talks after removing those sentences present in the parallel training data described above. For German, we used a corpus with 100,000 sentences randomly sampled from the News Commentary v16 German monolingual corpus.\footnote{\url{https://data.statmt.org/news-commentary/v16/training-monolingual/}}

Finally, we used the 
following corpora for the experiments in high-resource conditions. For English--German we used the training corpora available for the WMT 2014 shared task on machine translation;\footnote{\url{https://nlp.stanford.edu/projects/nmt/}} for development we used the concatenation of \emph{newstest2012} and \emph{newstest2013}, and for testing \emph{newstest2014}. For English--Romanian, we used the corpora available for the WMT 2016 shared task on machine translation of news; for training we used the concatenation of Europarl v8 and SETIMES2; for development and testing we used \emph{newsdev2016} and \emph{newstest2016}, respectively. Table \ref{tab:corpus} provides the amount of sentences and tokens in the training corpora used in our experiments. 


\begin{table}[tb]
\centering
\begin{tabular}{lrrr}
\textbf{Pair} & \textbf{\# sent.} & \textbf{\# left tok.} & \textbf{\#  right tok.} \\
\hline \hline
\rowcolor{gray!10}
\multicolumn{4}{c}{Low-resource conditions} \\
\hline
\multicolumn{4}{c}{Parallel data} \\
\hline
en--de & 174,443 & 3,575,407 & 3,353,855 \\
en--he & 187,817 & 3,862,985 & 2,958,136 \\
en--vi & 133,317 & 2,965,962 & 3,361,789 \\
de--hsb& 147,521 & 2,240,126 & 1,998,047 \\
de--rm & 102,192 & 1,773,683 & 2,414,749 \\
\hline
\multicolumn{4}{c}{Parallel data + back-translated data} \\
\hline
de--en & 269,213 & 5,537,986 & 5,843,264 \\
he--en & 282,587 & 4,728,840 & 6,130,842 \\
vi--en & 228,087 & 6,232,006 & 5,413,428 \\
hsb--de& 247,521 & 4,597,777 & 4.644.705 \\
rm--de & 202,192 & 5,412,126 & 4,178,262 \\
\hline\hline
\rowcolor{gray!10}
\multicolumn{4}{c}{High-resource conditions} \\
\hline
\multicolumn{4}{c}{Parallel data} \\
\hline
en--ro & 612,422 & 15,919,293 & 16,149,695 \\
en--de & 4,468,840 & 126,720,053 & 119,907,183 \\
\hline
\end{tabular}
\caption{Number of sentences and tokens in the training corpora used in our experiments. The first ten rows correspond to the experiments in low-resource conditions, only with parallel data and with parallel and back-translated data; the last two rows correspond to the  experiment in high-resource conditions (using only parallel data).}
\label{tab:corpus}
\end{table}

In order to study the domain robustness of 
MaTiLDA, 
we evaluated the systems trained for German--English and German--Romansh on out-of-domain test sets commonly used for this task \cite{muller-etal-2020-domain}.\footnote{\url{https://github.com/ZurichNLP/domain-robustness}} In line with Wang \& Sennrich~\cite{wang-sennrich-2020-exposure}, for German--English we chose IT, law and medical test sets, and for German--Romansh we chose a blog test set.


All corpora were tokenized and truecased with the Moses scripts;\footnote{\url{https://github.com/moses-smt/mosesdecoder/tree/master/scripts}} then, sentences longer than 100 tokens or with less than 5 tokens were removed from the training corpora. Afterwards, byte-pair encoding~\cite{sennrich-etal-2016-neural} (BPE) with 10,000 merge operations was applied on the concatenation of the source and target sides of the training corpora to obtain the vocabulary. Finally, those sentence pairs in the training corpora with more than 100 BPE tokens were removed.

One-to-many word alignments in both translation directions
were obtained using \texttt{mgiza++}~\cite{gao-vogel-2008-parallel}.\footnote{\url{https://github.com/moses-smt/mgiza}} Source-to-target word alignments were used for the \emph{mono} transformations \miqueladd{(see Sec.~\ref{sec:da})}; the one-to-one word alignments required by the \emph{replace} transformation were obtained by computing the intersection between the one-to-many word alignments in both translation directions.
The bilingual lexicon for the \emph{replace} transformation was built by  \miqueladd{annotating} each source word \miqueladd{with} the target word it is most frequently aligned with in the one-to-one word alignments.

\begin{table*}[tb]
\centering
\setlength{\tabcolsep}{4.5pt}
\begin{tabular}{lcccccccccccccccc}
\textbf{Task} & \textbf{en--de} & \textbf{de--en} & \textbf{en--he} & \textbf{he--en} & \textbf{en--vi} & \textbf{vi--en} & \textbf{de--hsb} & \textbf{hsb--de}& \textbf{de--rm} & \textbf{rm--de}\\
\hline\hline
baseline & $25.4\pm .2$ & $29.4\pm .1$ & $21.4\pm .3$ & $31.7\pm .6$ & $28.3\pm .4$ & $27.8\pm .5$ & $49.7\pm .5$ & $49.1\pm \phantom{0}.4$ & $49.3\pm .5$ & $50.6\pm .3$ \\
\arrayrulecolor{lightgray} \hline \arrayrulecolor{black} 
SwOut & $25.0\pm .4$ & $30.4\pm .3$ & $22.0\pm .0$ & $32.4\pm .6$ & $28.9\pm .4$ & $27.8\pm .6$ & $50.1\pm .8$ & $\mathbf{49.6\pm1.1}$ & $50.9\pm \phantom{0}.2$ & $51.7\pm \phantom{0}.3$ \\
RAML & $25.4\pm .3$ & $30.4\pm .2$ & $21.8\pm .6$ & $33.0\pm .4$ & $29.0\pm .3$ & $28.0\pm .1$ & $50.6\pm .3$ & $49.1\pm1.0$ & $49.7\pm1.9$ & $51.0\pm1.7$ \\
SwOut+RAML & $25.6\pm .4$ & $30.2\pm .4$ & $22.2\pm .2$ & $32.5\pm .6$ & $28.9\pm .5$ & $28.0\pm .8$ & $50.6\pm .9$ & $49.0\pm1.3$ & $50.9\pm1.1$ & $50.8\pm1.5$ \\
SeqMix & $25.9\pm .3$ & $30.6\pm .1$ & $\mathbf{22.5\pm .4}$ & $32.9\pm .6$ & $29.7\pm .3$ & $\mathbf{28.6\pm .3}$ & $50.1\pm .1$ & $48.7\pm1.1$ & $50.8\pm1.0$ & $49.6\pm \phantom{0}.3$ \\
\hline
swap & $25.4\pm .5$ & $30.5\pm .3$ & $\mathbf{22.2\pm .5}$ & $32.7\pm .5$ & $29.3\pm .2$ & $27.9\pm .8$ & $50.6\pm .3$ & $\mathbf{50.2\pm \phantom{0}.2}$ & $52.0\pm \phantom{0}.7$ & $52.4\pm \phantom{0}.5$ \\
unk & $25.5\pm .3$ & $30.4\pm .3$ & $21.6\pm .7$ & $32.7\pm .5$ & $29.4\pm .3$ & $28.3\pm .6$ & $50.5\pm .3$ & $49.9\pm \phantom{0}.2$ & $52.4\pm \phantom{0}.6$ & $52.2\pm \phantom{0}.4$ \\
source & $25.3\pm .0$ & $30.0\pm .3$ & $21.8\pm .3$ & $31.8\pm .5$ & $28.8\pm .4$ & $27.6\pm .5$ & $50.1\pm .2$ & $48.4\pm \phantom{0}.8$ & $51.9\pm \phantom{0}.4$ & $51.3\pm \phantom{0}.9$ \\
reverse & $\mathbf{25.9\pm .5}$ & $30.3\pm .1$ & $21.8\pm .3$ & $33.5\pm .3$ & $28.8\pm .4$ & $28.4\pm .4$ & $50.9\pm .5$ & $49.5\pm \phantom{0}.2$ & $51.8\pm \phantom{0}.3$ & $51.7\pm \phantom{0}.9$ \\
mono & $25.2\pm .2$ & $30.0\pm .3$ & $21.7\pm .5$ & $32.5\pm .6$ & $29.0\pm .2$ & $27.7\pm .1$ & $50.5\pm .7$ & $49.5\pm \phantom{0}.3$ & $51.9\pm \phantom{0}.6$ & $51.4\pm \phantom{0}.7$ \\
replace & $\mathbf{26.1\pm .6}$ & $\mathbf{31.5\pm .2}$ & $\mathbf{22.6\pm .5}$ & $33.5\pm .3$ & $\mathbf{30.1\pm .2}$ & $\mathbf{28.9\pm .6}$ & $\mathbf{51.1\pm .4}$ & $\mathbf{49.7\pm \phantom{0}.5}$ & $\mathbf{53.3\pm \phantom{0}.8}$ & $52.7\pm1.4$ \\
rev+repl & $\mathbf{26.3\pm .2}$ & $\mathbf{31.7\pm .6}$ & $\mathbf{22.9\pm .6}$ & $\mathbf{34.1\pm .5}$ & $\mathbf{30.1\pm .5}$ & $\mathbf{28.8\pm .4}$ & $51.1\pm .2$ & $\mathbf{49.9\pm \phantom{0}.4}$ & $\mathbf{53.0\pm \phantom{0}.7}$ & $52.8\pm1.0$ \\ 
rev+sw+repl & $\mathbf{26.5\pm .3}$ & $\mathbf{31.8\pm .4}$ & $\mathbf{22.8\pm .1}$ & $\mathbf{33.9\pm .6}$ & $\mathbf{30.4\pm .5}$ & $\mathbf{29.1\pm .6}$ & $\mathbf{51.6\pm .4}$ & $\mathbf{49.7\pm \phantom{0}.8}$ & $\mathbf{53.4\pm \phantom{0}.6}$ & $\mathbf{54.1\pm \phantom{0}.2}$ \\
\hline
\end{tabular}
\caption{For low-resource conditions, mean and standard deviation of the BLEU scores obtained when translating in-domain test sets with the baseline system, four other DA reference systems, and MaTiLDA, using different transformations and combinations of them. 
}
\label{tab:results-train-bleu}
\end{table*}

\subsection{Training}
Our neural model is a transformer \emph{base} model 
as defined by Vaswani et al.~\cite{vaswani2017attention}, with the exception of the amount of warm-up steps, which was set to 8,000. All the experiments were carried out on a single GPU with mini-batches made of 4,000 tokens. Validation was done every 1,000 updates in the low-resource scenario and every 5,000 updates in the high-resource one, and the patience, based on the BLEU score on the development set, was set to 6 validation cycles; we then kept the intermediate model performing best on the development set.
We trained the systems with the \texttt{fairseq} toolkit~\cite{ott-etal-2019-fairseq}. For RAML and SwitchOut, we integrated into \texttt{fairseq} the sampling function released by their  authors~\cite{wang-etal-2018-switchout}. For SeqMix, we also integrated the modifications in data processing and training loss published by their authors as a \texttt{fairseq} task.\footnote{Code available at \url{https://github.com/transducens/MaTiLDA}.}

Systems trained with MaTiLDA were fine-tuned on the original training samples after being trained on the combination of original and synthetic training samples. 
The results reported are those obtained with the model that maximizes BLEU on the development set.

As regards the DA hyperparameters, the proportion of words affected by the \emph{swap}, \emph{unk} and \emph{replace} transformations is controlled by a hyperparameter $\alpha$ for which we explored, for each translation task, values in $[0.1,0.9]$ at intervals of $0.1$. 
RAML and SwitchOut are governed by a temperature $\tau$. \miqueladd{To set its value we tried, for each translation task, a set of values around the best ones reported}
by Wang et al.~\cite{wang-etal-2018-switchout}.\footnote{$\tau^{-1} \in \{ 0.5, 0.6, 0.7, 0.8, 0.85, 0.9, 0.95, 1.0, 1.1, 1.2, 1.3 \}$} For the combination of SwitchOut and RAML, firstly the best $\tau_x$ for SwitchOut was determined and, afterwards, the best $\tau_y$ for RAML was sought by fixing $\tau_x$, as done by Wang et al. 
SeqMix is influenced by a hyperparameter that controls the sampling of the sub-parts of the training samples that are mixed up. We experimented with values in the interval $[0.1,1.5]$, similarly to Guo~et~al.~\cite{guo-etal-2020-sequence},\footnote{ $\{ 0.1, 0.5, 1.0, 1.5 \}$.  } and selected the systems performing best on the development set.

\section{Results and discussion}
\label{sec:results}
In this section we report the results achieved by MaTiLDA when it is used to train NMT systems for low-resource translation tasks (Sec.~\ref{se:low-resource}), when it is applied  in combination with
back-translation (Sec.~\ref{se:combine-backtranslation}), and when it is used in high-resource conditions (Sec.~\ref{se:high-resource}). We also report the results when the NMT systems 
are evaluated on out-of-domain test sets~(Sec.~\ref{se:domain-robust}). 

\subsection{Low-resource conditions}\label{se:low-resource}
Table \ref{tab:results-train-bleu} reports the mean and standard deviation of the translation performance, measured in terms of BLEU~\cite{papineni},\footnote{\texttt{sacrebleu}~\cite{post-2018-call} version string: \texttt{BLEU+case.mixed+lang.vi-en+numrefs.1+} \texttt{smooth.exp+tok.13a+version.1.5.0}} of three different executions \revadd{(with different random seeds)} for each of the systems built in low-resource conditions. 
\revadd{We employ the \emph{almost stochastic order} (ASO) method~\cite{dror2019deep}, following Ulmer et al.'s implementation~\cite{ulmer2022deep}, to assess statistical significance at a $p$-value threshold of 0.05. For every language pair, we highlight the scores achieved by the top-performing model in bold, as well as those 
whose difference with the top-performing model is not statistically significant.\footnote{\revadd{Following Ulmer et al.~\cite{ulmer2022deep}, we set an ASO decision threshold of~$\tau = 0.2$.}} 
These notation is used for the rest of results reported from this point onward. COMET and chrF++} scores, which were also computed, show a similar trend in all the experiments in this paper and will not be reported.


The results show that MaTiLDA consistently outperforms the baseline system  in all language pairs and translation directions, regardless of the transformations applied, except for \emph{source} (copying the source sentence; see results for vi--en and hsb--de) and \emph{mono} (reordering target words for monotonous alignment; see results for en--de and vi--en). In general, the transformations
\emph{replace} (random replacement of target words and the source words they are aligned with) and \emph{reverse} (translation into the target language but in the reverse order) are the best-performing ones; although for some language pairs \emph{swap} (random swapping of target words) behaves better than \emph{reverse}. 

Interestingly, training on synthetic training samples generated with the three best transformations  (\emph{reverse}+\emph{replace}+\emph{swap}) further improves the performance, achieving the best results in all translation tasks, except for en--he and he--en, for which the best results are obtained by combining only the two best transformations (\emph{reverse}+\emph{replace}), and for hsb--de for which the best results are obtained with a single transformation (\emph{swap}). This suggests that different transformations affect the NMT system in different ways and are somehow complementary. The improvement over the baseline in terms of BLEU  varies from 1.1 (hsb--de) to 4.1 (de--rm) BLEU points, being 2.1 BLEU points the average improvement. 

A comparison of MaTiLDA with RAML, SwitchOut, their combination (SwOut+RAML) and SeqMix, shows that our approach outperforms all of them. In general, of these four \miqueladd{reference systems}, the best performing one is SeqMix, which outperforms the others in six of the ten translation tasks. In any case, for all language pairs, MaTiLDA outperforms the best of these four reference systems in all cases, with improvements that range from 0.4 to 2.5 BLEU points; 1.1 BLEU points on average.

\begin{table}[tb]
\centering
\setlength{\tabcolsep}{4pt}
\begin{tabular}{lccccc}
\textbf{Task} & \textbf{de-en} & \textbf{he-en} & \textbf{vi-en} & \textbf{hsb-de} & \textbf{rm-de}\\
\hline \hline
baseline &	$31.4\pm.2$ & $34.0\pm.5$ & $29.4\pm.4$ & $49.4\pm.5$ & $49.1\pm.9$\\ 
MaTiLDA 
 &  $\mathbf{32.8\pm.1}$ & $\mathbf{34.8\pm.3}$ & $\mathbf{30.6\pm.4}$ & $\mathbf{50.3\pm.3}$ & $\mathbf{51.3\pm.5}$\\
\hline
\end{tabular}
\caption{For low-resource conditions and back-translated data, mean and standard deviation of the BLEU scores obtained when translating in-domain test sets with the baseline system, and MaTiLDA using the combination of the best transformations according to the experiments without back-translated data (\emph{reverse}+\emph{swap}+\emph{replace}).}
\label{tab:results-backtranslation}
\end{table}

\subsection{Combination with back-translation}\label{se:combine-backtranslation}
Next we explore the combination of the standard DA method, back-translation \cite{sennrich-back}, and MaTiLDA when the best performing transformations are used together (\emph{reverse}+\emph{replace}+\emph{swap}) \miqueladd{under the aforementioned low-resource conditions}. We do this only for the translation into the high resource languages, for which large monolingual corpora are available. The back-translated data was obtained by translating, with the baseline system used in the experiments reported in Table \ref{tab:results-train-bleu}, the corpora described in Sec.~\ref{se:datasets} from English into German, Hebrew and Vietnamese, and from German into Upper Sorbian and Romansh. The systems were then trained as usual on a corpus made of the original training corpus plus the synthetic corpus made of back-translated data (see Table \ref{tab:corpus}) on the source and the original English or German sentence (depending on the language pair) on the target. Note that MaTiLDA was applied to the back-translated data as well, just as if it were the original training corpus.\footnote{We tried different combinations of using the back-translated data: with and without applying MaTiLDA on back-translated sentence pairs, using a tag (as with the transformations) to flag back-translated sentence pairs, and using a tag to flag back-translated sentence pairs on which MaTiLDA was applied. All of them performed similarly.}

Table~\ref{tab:results-backtranslation} shows the results of the aforementioned experiment using back-translated data. A comparison of the performance, reported in tables \ref{tab:results-train-bleu} and \ref{tab:results-backtranslation} for the baseline systems, shows that when the target language is English, the use of back-translated data improves performance in around 2.0 BLEU points; when the target language is German the use of back-translation makes the resulting MT system to perform comparably (see results for hsb--de) or even worse (see results for rm--de). This difference between the systems translating into English and into German may be explained by the fact that, for hsb--de and rm--de, the original training corpus and the test set come from the same source (hsb--de; Witaj Language Center) or belong to the same domain (rm--de; law) ---note the high BLEU scores for these language pairs as compared to the others--- whereas the German monolingual corpus used to generate the back-translated data comes from a different source (News Commentary) and belongs to a different domain (news). 

A comparison of the results obtained when back-translation and MaTiLDA are used together shows an improvement over the system using only back-translation of around 1.3 BLEU points, which accounts for the complementarity of both DA approaches. Compared to the application of MaTiLDA alone (last row of Table~\ref{tab:results-train-bleu}) the improvement of the systems translating into English is around 1.1 BLEU points. When translating into German, translation performance is improved by 0.6 BLEU points in the case of hsb--de, and worsened by 2.8 BLEU points in the case of rm--de. Note that the rm--de baseline system trained on back-translated data performs worse than the baseline system trained solely on parallel corpora (1.5 BLEU points worse).

\begin{table}
\centering
\setlength{\tabcolsep}{4pt}
\begin{tabular}{lcccccccccccccc}
\textbf{Task} & \textbf{en--ro} & \textbf{ro--en} & \textbf{en--de} & \textbf{de--en}  \\
\hline \hline
baseline & $23.3\pm.1$ & $30.5\pm.2$ & $24.3\pm.5$ & $30.0\pm.4$   \\ 
\arrayrulecolor{lightgray} \hline \arrayrulecolor{black} 
SwitchOut & $23.3\pm.3$ & $30.9\pm.2$ & $24.8\pm.2$ &  $30.1\pm.8$  \\ 
RAML & $23.6\pm.2$ &  $30.7\pm.1$ & $24.9\pm.1$ & $30.6\pm.6$  \\ 
SwOut+RAML & $23.5\pm.2$ & $31.2\pm.1$ & $24.7\pm.2$ &  $30.0\pm.3$ \\ 
SeqMix  & $23.5\pm.2$ & $31.3\pm.3$ & $24.6\pm.2$ &  $30.0\pm.8$ \\ 
\hline
swap & $23.5\pm.2$ & $30.9\pm.4$ & $24.9\pm.4$& $\mathbf{30.7\pm.7}$\\ 
unk & $23.6\pm.1$ & $30.7\pm.0$ & $25.3\pm.2$ & $\mathbf{31.1\pm.2}$\\ 
source & $23.6\pm.2$ & $31.1\pm.3$ & $23.9\pm.6$ & $29.7\pm.5$   \\ 
reverse & $23.8\pm.1$ & $31.2\pm.2$ & $24.3\pm.3$ & $30.2\pm.7$ \\ 
mono & $23.3\pm.4$ & $30.4\pm.2$ & $24.6\pm.2$ & $30.5\pm.2$   \\ 
replace & $23.9 \pm .2$ & $31.6 \pm .3$ & $\mathbf{25.6\pm.2}$ & $\mathbf{30.9\pm.1}$  \\ 
sw+unk+repl & $\mathbf{24.3\pm.2}$ & $\mathbf{32.1\pm.1}$ & $\mathbf{25.8\pm.2}$& $\mathbf{31.5\pm.8}$  \\
\hline
\end{tabular}
\caption{For high-resource conditions, mean and standard deviation of the BLEU scores obtained when translating in-domain test sets with the baseline system, four other DA reference systems, and MaTiLDA, using different transformations and combinations of them. 
}
\label{tab:results-train-bleu-high}
\end{table}

\subsection{High-resource conditions}\label{se:high-resource}
Even thought MaTiLDA is aimed at improving the translation quality of NMT systems trained on scarce parallel corpora, \miqueladd{we have also studied} the performance of MaTiLDA when it is used to train systems in high-resource conditions. Table~\ref{tab:results-train-bleu-high} reports the results of these experiments for 
English--Romanian and English--German in both translation directions.
For English--Romanian the training corpus is around 4 times larger than for the low-resource languages with which we have experimented so far; for English--German the training corpus is around 30 times larger (see Table~\ref{tab:corpus}).

\begin{table*}[tb]
\centering
\setlength{\tabcolsep}{4.5pt}
\begin{tabular}{lrrrrrrrrrrrrrr}
\textbf{Domain} & \multicolumn{2}{c}{IT} & \multicolumn{2}{c}{Law} & \multicolumn{2}{c}{Medical} & \multicolumn{2}{c}{Blogs} \\
\hline
\textbf{Direction} & \multicolumn{1}{c}{\textbf{en--de}} & \multicolumn{1}{c}{\textbf{de--en}} & \multicolumn{1}{c}{\textbf{en--de}} & \multicolumn{1}{c}{\textbf{de--en}} & \multicolumn{1}{c}{\textbf{en--de}} & \multicolumn{1}{c}{\textbf{de--en}} & \multicolumn{1}{c}{\textbf{de--rm}} & \multicolumn{1}{c}{\textbf{rm--de}} \\
\hline 
\hline
baseline & $6.9\pm\phantom{0}.4$ & $4.5\pm1.8$ & $7.7\pm\phantom{0}.5$ & $7.6\pm1.8$ & $11.0\pm.9$ & $10.0\pm2.0$ & $15.8\pm.2$ & $15.2\pm.3$ \\
\arrayrulecolor{lightgray} \hline \arrayrulecolor{black} 
SwOut+RAML & $5.9\pm1.4$ & $8.1\pm1.4$ & $7.0\pm1.1$ & $7.8\pm\phantom{0}.8$ & $12.0\pm.4$ & $11.3\pm1.2$ & $16.1\pm.7$ & $15.4\pm.4$ \\
SeqMix & $8.3\pm1.8$ & $9.4\pm1.3$ & $8.7\pm\phantom{0}.4$ & $9.3\pm\phantom{0}.2$ & $12.7\pm.9$ & $12.0\pm\phantom{0}.4$ & $17.5\pm.3$ & $16.3\pm.4$ \\
MaTiLDA & $\mathbf{15.3\pm\phantom{0}.3}$ & $\mathbf{13.3\pm\phantom{0}.7}$ & $\mathbf{10.2\pm\phantom{0}.5}$ & $\mathbf{10.7\pm\phantom{0}.1}$ & $\mathbf{16.9\pm.6}$ & $\mathbf{15.1\pm1.0}$ & $\mathbf{20.8\pm.1}$ & $\mathbf{19.8\pm.5}$ \\ 
\hline
\end{tabular}
\caption{For low-resource conditions, mean and standard deviation of the BLEU scores obtained when translating out-of-domain test sets. \miqueladd{The MaTiLDA results were computed using the combination of the best transformations on low-resource conditions (\emph{reverse}+\emph{swap}+\emph{replace})}.}
\label{tab:results-outdomain-bleu}
\end{table*}

\begin{table*}[tb]
\centering
\setlength{\tabcolsep}{4.5pt}
\begin{tabular}{lrrrrrrrrrrrrrr}
\textbf{Domain} & \multicolumn{2}{c}{IT} & \multicolumn{2}{c}{Law} & \multicolumn{2}{c}{Medical}  \\
\hline
\textbf{Direction} & \multicolumn{1}{c}{\textbf{en--de}} & \multicolumn{1}{c}{\textbf{de--en}} & \multicolumn{1}{c}{\textbf{en--de}} & \multicolumn{1}{c}{\textbf{de--en}} & \multicolumn{1}{c}{\textbf{en--de}} & \multicolumn{1}{c}{\textbf{de--en}}  \\
\hline 
\hline
baseline & $\mathbf{10.6\pm2.1}$ & $20.8\pm1.0$ & $28.3\pm2.4$ & $32.0\pm \phantom{0}.1$ & $17.4\pm1.0$ & $23.3\pm .9$ \\
\arrayrulecolor{lightgray} \hline \arrayrulecolor{black} 
SwOut+RAML & $\mathbf{11.5\pm3.2}$ & $20.8\pm1.8$ & $28.9\pm1.4$ & $33.7\pm \phantom{0}.2$ & $18.3\pm1.4$ & $25.0\pm .4$ \\
SeqMix & $\mathbf{11.2\pm2.0}$ & $18.5\pm2.6$ & $28.1\pm1.1$ & $30.1\pm1.4$ & $17.7\pm1.0$ & $23.3\pm .7$ \\
MaTiLDA &$\mathbf{13.7\pm2.7}$ & $\mathbf{29.0\pm1.1}$ & $\mathbf{31.0\pm1.9}$ & $\mathbf{35.7\pm \phantom{0}.8}$ & $\mathbf{22.2\pm1.4}$ & $\mathbf{29.1\pm .8}$ \\ 
\hline
\end{tabular}
\caption{For high-resource conditions, mean and standard deviation of the BLEU scores obtained when translating out-of-domain test sets. \miqueladd{The MaTiLDA results were computed using the combination of the best transformations on high-resource conditions (\emph{swap}+\emph{unk}+\emph{replace})}.}
\label{tab:results-outdomain-bleu-high}
\end{table*}

As the results in Table \ref{tab:results-train-bleu-high} show, MaTiLDA outperforms the baseline in the four translation tasks \miqueladd{---}the improvement ranges from 1.0 to 1.6 BLEU points; 1.4 BLEU points on average\miqueladd{--- as well as the other DA approaches}. It is worth noting that 
the improvement it brings seems not to be conditioned by the amount of parallel data used, as it is  around 1.3 BLEU point for English--Romanian (about 600,000 training parallel sentences) and 1.5 BLEU points for English--German (more than 4.4 million parallel sentences). As happened with the low-resource translation tasks, almost all transformations improve over the baseline, and the combination of the best performing ones (\emph{swap}+\emph{unk}+\emph{replace}) further improve the baseline results. 

\subsection{Domain robustness} \label{se:domain-robust}
Finally, we evaluate the performance of the English--German and German--Romansh NMT systems when translating out-of-domain test sets.
For English--German we used test sets in the IT, law, and medical domains; for German--Romansh we used a test set whose sentence pairs were extracted from blogs  (see Sec.~\ref{se:datasets} for more details). 

Tables \ref{tab:results-outdomain-bleu} and \ref{tab:results-outdomain-bleu-high} report the translation performance attained by  the NMT systems trained in low-resource and high-resource conditions, respectively, when they are evaluated on out-of-domain test sets. The systems being evaluated are: the baseline system, MaTiLDA using the best three transformations, the combination of SwitchOut and RAML (SwOut+RAML) and SeqMix.
These tables show that MaTiLDA outperforms the baseline, SwOut+RAML and SeqMix systems both in low-resource and high-resource conditions. The best performing \miqueladd{reference} system in low-resource conditions is SeqMix and MaTiLDA outperforms SeqMix by 2.7 BLEU points on average. In high-resource conditions the best performing \miqueladd{reference} system is the combination of SwitchOut and RAML (SwOut+RAML) and MaTiLDA outperforms SwOut+RAML by 3.8 BLEU points on average. Note that there are cases, such as the IT domain for German-English, in which the improvement is above 8 BLEU points.

\section{Explainability}
\label{sec:explain}

\begin{table*}[tb]
\centering
\setlength{\tabcolsep}{3.1pt}
\begin{tabular}{lcccccccccccccccc}
\textbf{Task} & \textbf{en--de} & \textbf{de--en} & \textbf{en--he} & \textbf{he--en} & \textbf{en--vi} & \textbf{vi--en} & \textbf{de--hsb} & \textbf{hsb--de}& \textbf{de--rm} & \textbf{rm--de}\\
\hline \hline
baseline & $63.2\pm \phantom{0}.8$ & $67.1\pm2.0$ &  $71.8\pm1.7$ & $67.2\pm2.6$ & $68.1\pm1.1$ & $58.7\pm1.2$ & $77.6\pm1.1$ & $76.2\pm1.4$ & $72.9\pm \phantom{0}.9$ & $63.8\pm1.7$ \\
\arrayrulecolor{lightgray} \hline \arrayrulecolor{black} 
SwOut & $60.4\pm \phantom{0}.2$ & $63.1\pm1.1$  & $68.5\pm \phantom{0}.7$ & $64.2\pm3.5$ & $64.2\pm1.1$ & $56.7\pm \phantom{0}.2$ & $72.2\pm \phantom{0}.8$ & $67.1\pm1.1$ & $69.0\pm \phantom{0}.7$ & $58.8\pm1.9$ \\
RAML & $\mathbf{68.7\pm1.8}$ & $70.3\pm1.7$ & $74.3\pm2.2$ & $72.2\pm2.2$ & $73.2\pm3.0$ & $60.9\pm1.4$ & $79.6\pm1.5$ & $79.9\pm1.2$ & $75.8\pm1.4$ & $69.3\pm1.9$ \\
SwOut+RAML & $65.4\pm2.5$ & $68.5\pm1.9$ & $71.0\pm \phantom{0}.8$ & $67.0\pm1.5$ & $67.5\pm2.3$ & $59.1\pm \phantom{0}.6$ & $76.0\pm1.8$ &$72.5\pm3.1$ & $73.6\pm2.2$ & $64.8\pm1.8$ \\
SeqMix & $63.5\pm \phantom{0}.6$ & $64.2\pm2.1$ & $67.8\pm1.5$ & $62.8\pm1.2$ & $58.0\pm1.6$ & $54.7\pm \phantom{0}.9$ & $75.0\pm2.8$ & $75.2\pm1.5$ & $57.6\pm1.6$ & $63.5\pm1.4$ \\
\hline
swap & $68.3\pm \phantom{0}.7$ & $\mathbf{74.9\pm2.3}$ & $76.9\pm \phantom{0}.8$ & $74.0\pm1.6$ & $74.0\pm1.3$ & $63.4\pm1.9$ & $79.7\pm1.4$ & $78.9\pm1.2$ & $\mathbf{79.6\pm \phantom{0}.3}$ & $68.4\pm1.8$ \\ 
unk & $\mathbf{71.2\pm \phantom{0}.8}$ & $74.3\pm \phantom{0}.7$ & $75.8\pm1.2$ & $71.2\pm1.6$ & $71.5\pm2.0$ & $\mathbf{70.3\pm1.2}$ & $\mathbf{84.1\pm1.1}$ & $\mathbf{84.6\pm1.4}$ & $\mathbf{80.2\pm1.2}$ & $66.1\pm \phantom{0}.7$ \\ 
source & $68.6\pm1.5$ & $71.8\pm1.8$ & $74.6\pm1.6$ & $67.8\pm2.9$ & $71.1\pm1.3$ & $64.3\pm \phantom{0}.4$ & $82.0\pm \phantom{0}.6$ & $81.6\pm2.2$ & $76.8\pm \phantom{0}.0$ & $68.3\pm1.8$ \\ 
reverse & $\mathbf{71.3\pm2.2}$ & $\mathbf{75.1\pm \phantom{0}.5}$ & $\mathbf{80.1\pm1.0}$ & $\mathbf{76.3\pm1.5}$ & $\mathbf{77.2\pm \phantom{0}.9}$ & $\mathbf{70.6\pm2.1}$ & $\mathbf{84.0\pm1.6}$ & $\mathbf{83.0\pm \phantom{0}.9}$ & $\mathbf{80.0\pm1.5}$ & $\mathbf{75.1\pm1.2}$ \\ 
mono & $64.7\pm1.7$ & $70.0\pm \phantom{0}.3$ & $72.4\pm \phantom{0}.1$ & $66.9\pm1.1$ & $69.6\pm1.5$ & $61.7\pm2.0$ & $76.4\pm \phantom{0}.7$ & $76.6\pm \phantom{0}.6$ & $75.5\pm \phantom{0}.9$ & $66.7\pm1.7$ \\ 
replace & $62.1\pm1.9$ & $67.1\pm \phantom{0}.2$ & $71.6\pm1.6$ & $72.5\pm1.4$ & $68.3\pm1.0$ & $60.5\pm1.6$ & $76.0\pm1.8$ & $75.9\pm2.3$ & $72.0\pm1.0$ & $65.2\pm1.0$ \\ 
rev+repl & $\mathbf{67.9\pm3.0}$ & $73.5\pm1.2$ & $78.9\pm1.2$ & $\mathbf{77.7\pm \phantom{0}.9}$ & $71.9\pm2.2$ & $65.7\pm1.4$ & $81.0\pm1.9$ & $\mathbf{83.6\pm1.5}$ & $76.1\pm \phantom{0}.5$ & $71.7\pm \phantom{0}.4$ \\ 
rev+sw+repl & $\mathbf{69.3\pm2.5}$ & $\mathbf{74.7\pm1.2}$ & $\mathbf{81.1\pm \phantom{0}.1}$ & $\mathbf{77.0\pm \phantom{0}.7}$ & $\mathbf{75.8\pm2.3}$ & $65.6\pm1.3$ & $81.5\pm1.7$ & $82.1\pm1.7$ & $\mathbf{79.4\pm1.1}$ & $71.9\pm2.8$ \\
\hline
\end{tabular}
\caption{For low-resource conditions, mean and standard deviation of the source influence obtained when translating in-domain test sets with the baseline system, four other DA reference systems, and MaTiLDA using different transformations and combinations of them.}
\label{tab:results-source-influence}
\end{table*}

\begin{table}[tb]
\centering
\setlength{\tabcolsep}{4pt}
\begin{tabular}{lcccccccccccccc}
\textbf{Task} & \textbf{en--ro} & \textbf{ro--en} & \textbf{en--de} & \textbf{de--en}  \\
\hline \hline
baseline & $72.0\pm \phantom{0}.9$ & $84.3\pm \phantom{0}.6$ & $\mathbf{62.5\pm5.3}$ & $68.4\pm1.0$ \\
\arrayrulecolor{lightgray} \hline \arrayrulecolor{black} 
SwitchOut &  $68.6\pm1.1$ & $80.8\pm \phantom{0}.9$ & $56.9\pm4.1$ & $67.7\pm \phantom{0}.9$ \\ 
RAML &  $\mathbf{78.5\pm \phantom{0}.6}$ & $\mathbf{87.7\pm1.6}$   & $61.9\pm \phantom{0}.6$ & $\mathbf{76.9\pm2.8}$ \\ 
SwOut+RAML &  $76.2\pm1.0$ & $86.1\pm \phantom{0}.5$  & $\mathbf{65.1\pm5.0}$ & $\mathbf{77.7\pm \phantom{0}.7}$ \\ 
SeqMix &  $71.6\pm1.2$ & $80.6\pm1.6$ & $56.5\pm2.1$ & $69.7\pm4.5$ \\
\hline
swap & $76.1\pm \phantom{0}.4$ & $\mathbf{86.9\pm \phantom{0}.7}$ & $\mathbf{61.8\pm4.3}$ & $72.4\pm \phantom{0}.2$ \\ 
unk & $\mathbf{76.8\pm2.4}$ & $85.9\pm \phantom{0}.3$ & $59.3\pm3.9$ & $72.7\pm1.2$ \\ 
source & $70.4\pm2.5$ & $82.0\pm \phantom{0}.5$ & $\mathbf{65.9\pm3.5}$ & $71.4\pm \phantom{0}.8$ \\ 
reverse & $71.0\pm1.2$ & $\mathbf{86.5\pm1.5}$ & $\mathbf{60.4\pm7.0}$ & $69.1\pm1.6$ \\ 
mono & $74.3\pm2.1$ & $\mathbf{87.3\pm1.1}$ & $\mathbf{66.5\pm1.9}$ & $72.7\pm1.7$ \\ 
replace & $70.7\pm \phantom{0}.8$ & $80.3\pm \phantom{0}.8$ & $\mathbf{66.0\pm4.6}$ & $\mathbf{75.7\pm2.2}$ \\ 
sw+unk+repl & $75.9\pm \phantom{0}.9$ & $79.7\pm1.1$ & $62.4\pm3.2$ & $74.6\pm1.4$ \\
\hline
\end{tabular}
\caption{For high-resource conditions, mean and standard deviation of the source influence obtained when translating in-domain test sets with the baseline system, four other DA reference systems, and MaTiLDA using different transformations and combinations of them. 
}
\label{tab:results-source-influence-high}
\end{table}

\begin{table*}[tb]
\centering
\setlength{\tabcolsep}{4.5pt}
\begin{tabular}{lcccccccccccccc}
\textbf{Domain} & \multicolumn{2}{c}{IT} & \multicolumn{2}{c}{Law} & \multicolumn{2}{c}{Medical} & \multicolumn{2}{c}{Blogs} \\
\hline
\textbf{Direction} & \textbf{en--de} & \textbf{de--en} & \textbf{en--de} & \textbf{de--en} & \textbf{en--de} & \textbf{de--en} & \textbf{de--rm} & \textbf{rm--de} \\
\hline 
\hline
baseline & $62.4\pm1.4$ & $66.2\pm1.9$ & $58.8\pm \phantom{0}.7$ & $58.7\pm1.6$ & $\mathbf{62.0\pm1.0}$ & $62.6\pm2.2$ & $64.8\pm \phantom{0}.9$ & $61.4\pm1.3$ \\
\arrayrulecolor{lightgray} \hline \arrayrulecolor{black} 
SwOut+RAML & $\mathbf{66.1\pm2.1}$ & $68.1\pm2.4$ & $\mathbf{62.4\pm2.6}$ & $61.0\pm1.9$ & $\mathbf{65.3\pm2.1}$ & $64.6\pm2.2$ & $67.2\pm3.2$ & $62.4\pm1.9$ \\
SeqMix & $61.6\pm1.7$ & $66.1\pm2.1$ & $60.2\pm1.4$ & $61.8\pm2.1$ & $\mathbf{62.4\pm1.1}$ & $63.4\pm2.2$ & $55.9\pm1.2$ & $54.8\pm1.2$ \\
MaTiLDA & $\mathbf{66.8\pm2.4}$ & $\mathbf{76.4\pm1.1}$ & $\mathbf{66.3\pm3.2}$ & $\mathbf{70.7\pm \phantom{0}.6}$ & $\mathbf{66.2\pm3.2}$ & $\mathbf{72.6\pm \phantom{0}.7}$ & $\mathbf{74.1\pm1.7}$ & $\mathbf{68.8\pm2.8}$ \\
\hline
\end{tabular}
\caption{For low-resource conditions, mean and standard deviation of the source influence obtained when translating out-of-domain texts in the IT, law, medical and blogs domains. \miqueladd{The MaTiLDA results were computed using the combination of the best transformations on low-resource conditions (\emph{reverse}+\emph{swap}+\emph{replace})}}
\label{tab:results-outdomain-sourceinfluence}
\end{table*}

\begin{table*}[tb]
\centering
\setlength{\tabcolsep}{4.5pt}
\begin{tabular}{lrrrrrrrrrrrrrr}
\textbf{Domain} & \multicolumn{2}{c}{IT} & \multicolumn{2}{c}{Law} & \multicolumn{2}{c}{Medical}  \\
\hline
\textbf{Direction} & \multicolumn{1}{c}{\textbf{en--de}} & \multicolumn{1}{c}{\textbf{de--en}} & \multicolumn{1}{c}{\textbf{en--de}} & \multicolumn{1}{c}{\textbf{de--en}} & \multicolumn{1}{c}{\textbf{en--de}} & \multicolumn{1}{c}{\textbf{de--en}}  \\
\hline 
\hline
baseline & $\mathbf{61.3\pm4.8}$ & $75.4\pm1.3$ & $\mathbf{58.1\pm6.1}$ & $71.2\pm1.3$ & $\mathbf{59.9\pm4.4}$ & $65.8\pm1.4$ \\
\arrayrulecolor{lightgray} \hline \arrayrulecolor{black} 
SwOut+RAML & $\mathbf{61.7\pm4.8}$ & $\mathbf{81.2\pm1.0}$ & $\mathbf{63.1\pm6.4}$ & $\mathbf{78.8\pm1.3}$ & $\mathbf{60.0\pm5.2}$ & $\mathbf{70.8\pm1.3}$ \\
SeqMix & $\mathbf{54.8\pm3.7}$ & $\mathbf{74.3\pm4.7}$ & $52.2\pm1.2$ & $71.8\pm5.3$ & $\mathbf{53.6\pm3.5}$ & $65.2\pm4.4$ \\
MaTiLDA & $\mathbf{60.7\pm1.1}$ & $\mathbf{81.6\pm2.8}$ & $\mathbf{57.9\pm3.3}$ & $\mathbf{79.4\pm2.3}$ & $\mathbf{59.2\pm1.2}$ & $\mathbf{72.4\pm1.9}$ \\ 
\hline
\end{tabular}
\caption{For high-resource conditions,  mean and standard deviation of the source influence obtained when translating out-of-domain test sets. \miqueladd{The MaTiLDA results were computed using the combination of the best transformations on high-resource conditions (\emph{swap}+\emph{unk}+\emph{replace})}.}
\label{tab:results-outdomain-sourceinfluence-high}
\end{table*}


In the previous section we have exhaustively evaluated the performance of MaTiLDA and have shown that it systematically outperforms state-of-the-art DA methods both when translating in-domain and out-of-domain test sets. In this section we study if the use of MaTiLDA increases the contribution of the source representations produced by the encoder to the generation decisions made by the decoder (Sec.~\ref{se:srctgtcontrib}). In addition, we also study if MaTiLDA \miqueladd{leads} NMT systems to produce less hallucinations~\cite{hallucinations2018}, i.e., completely inadequate output translations \miqueladd{which are }strongly unrelated to the input text (Sec.~\ref{se:hallucinations}).

\subsection{Relative source and target contributions}\label{se:srctgtcontrib}

To compute the relative contribution of source and target tokens to each prediction made by the system we used an embedding perturbation method~\cite{ferrando-costa-jussa-2021-attention-weights}. Given a source sentence $\mathbf{x}$ and its translation $\mathbf{y}$, the absolute source contribution $C_S(y_j)$ when producing the probability of the $j$-th token $y_j$ is \miqueladd{defined} as the variance of $y_j$'s output probability across $N$ random perturbations of the word embeddings of $\mathbf{x}$. Specifically, $C_S(y_j)$ is computed according to the following equation:
\begin{displaymath}
C_S(y_j) = \frac{1}{N} \sum_{n=1}^{N} \left(p(y_j|\mathbf{y_{<j}},\mathbf{\tilde{x}^n}) -   \frac{1}{N} \sum_{m=1}^{N} p(y_j|\mathbf{y_{<j}},\mathbf{\tilde{x}^{m}})\right)^2\mbox{,}
\end{displaymath}
where $\mathbf{\tilde{x}^k}$ stands for the $k$-th perturbation of the word embeddings of $\mathbf{x}$.

In order to perturb the word embedding of a source token $x$, Gaussian noise with a standard deviation proportional to the Euclidean norm of the embedding is added to it:\footnote{We set $N = 50$ and $\lambda =0.01$~\cite{ferrando-costa-jussa-2021-attention-weights}.}
\begin{displaymath}
\tilde{x} = x + \mathcal{N}(0, \sigma_{x}^2); \sigma_{x} = \lambda \cdot ||x|| 
\end{displaymath}

The absolute target contribution $C_T(y_j)$ is computed analogously by perturbing $\mathbf{y_{<j}}$ instead of $\mathbf{x}$. The relative source contribution $C_{\mathrm{SR}}(y_j)$, which we use in our analysis, is then obtained after normalizing $C_S(y_j)$ as follows:
\begin{displaymath}
    C_{\mathrm{SR}}(y_j) = \frac{C_S(y_j)}{C_S(y_j) + C_T(y_j)} \mbox{.}
\end{displaymath}

We analysed the values of $C_{\mathrm{SR}}(y_j)$ in two different ways to shed light on the way MaTiLDA \miqueladd{affects} reliance on the source language information. On the one hand, we averaged $C_{\mathrm{SR}}(y_j)$ for all the tokens of the translation of the test set produced by a system to obtain an estimation of its general degree of reliance on the source language. On the other hand, we studied how $C_{\mathrm{SR}}(y_j)$ changes throughout the different tokens of the target sentence for each of the DA approaches. \miqueladd{In both cases, and following Voita et~al.~\cite{voita2020analyzing}, we teacher force the reference translations so as to perform comparisons between systems when producing the same output.}

%
%
%
Tables~\ref{tab:results-source-influence} and \ref{tab:results-source-influence-high} show the general source influence for systems trained, respectively, in low-resource and high-resource conditions, when translating the in-domain test sets. \revadd{As before, we used the \emph{almost stochastic order} (ASO) method ~\cite{dror2019deep,ulmer2022deep} to determine if the variations in source influence among the assessed systems are statistically significant, using a $p$-value threshold of 0.5. We highlight in bold the system with the highest source influence, along with those that do not exhibit statistically significant differences from it; we do this in all the tables reporting source influences.}

\revadd{As tables~\ref{tab:results-source-influence} and \ref{tab:results-source-influence-high} show,} in low-resource conditions, MaTiLDA systematically increases reliance on the source language, 
\miqueladd{thus, behaving similarly to NMT systems trained} on larger original parallel corpora~\cite{voita2020analyzing}. \emph{Reverse} and \emph{unk} are the transformations that bring the largest increase, and \emph{replace} the one that brings the smallest one. 
\miqueladd{The combination of the best transformations also brings a large increase in source-language reliance.}
Concerning the other DA approaches, only RAML increases source reliance. Note that this approach involves replacing some words in target-language sentences with other words randomly chosen from the vocabulary. Thus, it makes the target prefix less predictive, similarly to the MaTiLDA transformations. Nevertheless, both BLEU scores and source reliance are lower than those obtained with MaTiLDA.

In high-resource conditions, the increase in source reliance brought by MaTiLDA is smaller than in low-resource conditions, and it is focused mainly on the systems translating into English. When English is the source language, the target language (German and Romanian) is highly inflected and may require more reliance on the target context to produce a grammatical output. 

The general source influence when translating the different out-of-domain test sets, depicted in 
tables~\ref{tab:results-outdomain-sourceinfluence} and \ref{tab:results-outdomain-sourceinfluence-high},  follows similar trends to those identified for the in-domain scenario. The largest source influence is achieved by MaTiLDA for low-resource translation tasks, while the gap between it and SwitchOut+RAML vanishes when larger training data is available.

\begin{figure}
\centering
         \includegraphics[width=0.5\textwidth]{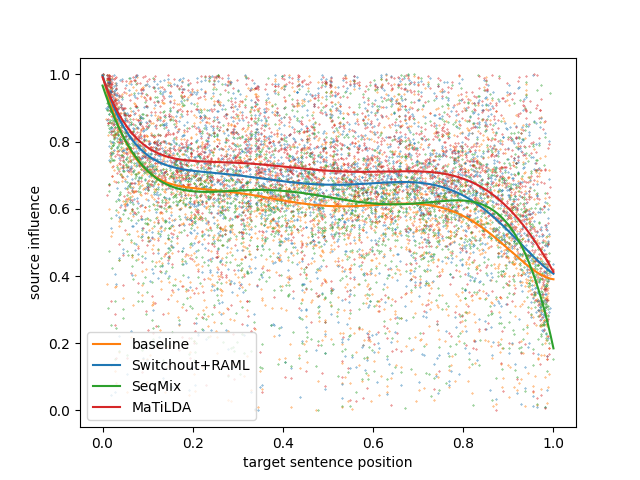}
        \caption{Source influence throughout relative target sentence positions for the English--German low-resource in-domain test set.}
        \label{fig:source-influence-over-time}
\end{figure}

Figure~\ref{fig:source-influence-over-time} depicts the value of $C_{\mathrm{SR}}(y_j)$ for each target-language token of the low-resource in-domain English--German test and for the different DA methods evaluated; the rest of language pairs show the same behaviour. As sentences have different lengths, the position of a token in the sentence (x-axis) is represented as the proportion of the words already predicted, hence the values in the x-axis are in the interval $[0,1]$, being $0$ the first token of the sentence and $1$ the last one. The polynomial that best fits the data (obtained via least squares) is also depicted in the plot.\footnote{\miqueladd{The degree of the polynomial was empirically obtained by incrementally exploring different values. We found that the lowest polynomial degree that best fits the data was 6.}} It can be observed that the source influence decreases as decoding progresses, in line with the findings of Voita et~al.~\cite{voita2020analyzing}. The difference between MaTiLDA and the baseline remains relatively constant through the sentence except for the first tokens, where the target prefix is too short to make a difference between the DA methods. The relative position between the DA methods evaluated matches those shown in Table~\ref{tab:results-source-influence}. 

\subsection{Hallucinations}\label{se:hallucinations}
In spite of the significant improvements in translation performance brought by the recent advancements in NMT, the phenomenon of hallucinations in NMT still remains a concern. 
Hallucinations usually appear in out-of-domain translations ~\cite{muller-sennrich-2021-understanding}, and may undermine user trust.

Hallucinations have been associated with systems failing to use source information properly~\cite{voita2020analyzing,detecting2022}, thus showing abnormal patterns in the cross-attention to the encoder. As we have already proved that MaTiLDA consistently improves source relevance, in this section we analyse whether this additionally results in \miqueladd{the} mitigation of hallucinations in both in-domain and out-of-domain translation tasks.

In order to estimate the number of hallucinations produced by systems trained with different DA techniques we used 
LaBSE~\cite{feng-etal-2022-labse} cross-lingual sentence embeddings. A number of studies~\cite{detecting2022,optimal2022,hallucinations2023lrp} have shown that the use of cross-lingual sentence embeddings to compute the similarity of MT outputs and their reference translations outperforms previous methods for detecting hallucinations, such as COMET~\cite{rei-etal-2020-comet,needle2022} or lexical-based metrics~\cite{hallucinations2018,muller-sennrich-2021-understanding}. 
\miqueladd{Noticeably, LaBSE, as a discriminator of hallucinations, unlike COMET, has been shown to differentiate hallucinations from poor translations~\cite{detecting2022}}.

We evaluate the tendency to hallucinate of \miqueladd{four systems:} the baseline, MaTiLDA including the best performing transformations, the combination of SwitchOut and RAML, and SeqMix. We compute the sentence-level LaBSE embeddings for the system outputs and the references, and then represent the cosine similarities between them. 
Computing the cosine similarity between the systems' output and the source sentences in the test sets results in similar plots. 

\begin{figure*}[tb]
\centering
     \begin{subfigure}[b]{0.24\textwidth}
         \centering
         \includegraphics[width=\textwidth]{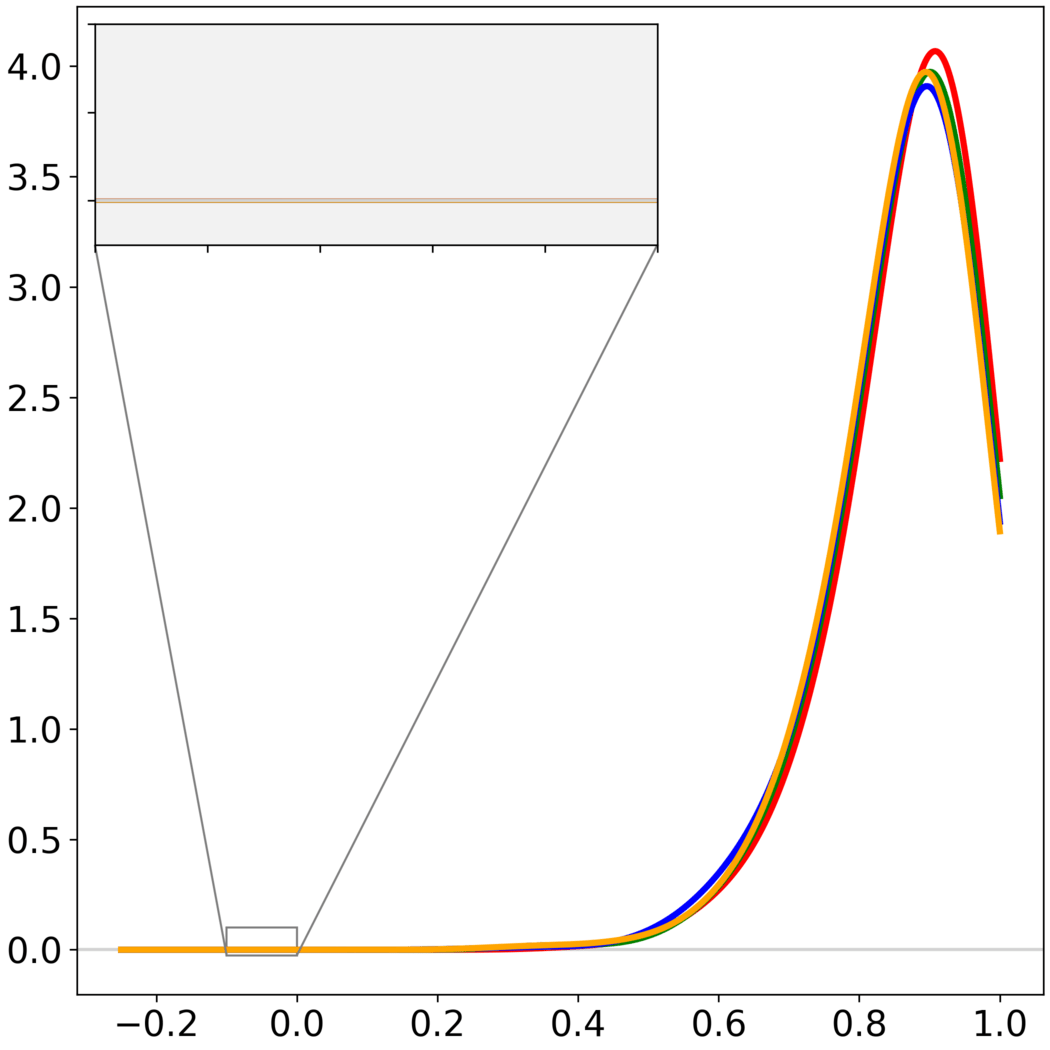}
         \caption{en--de, in-domain}
         \label{en-de-in-domain-kde-ref}
     \end{subfigure}
     \hfill
     \begin{subfigure}[b]{0.24\textwidth}
         \centering
         \includegraphics[width=\textwidth]{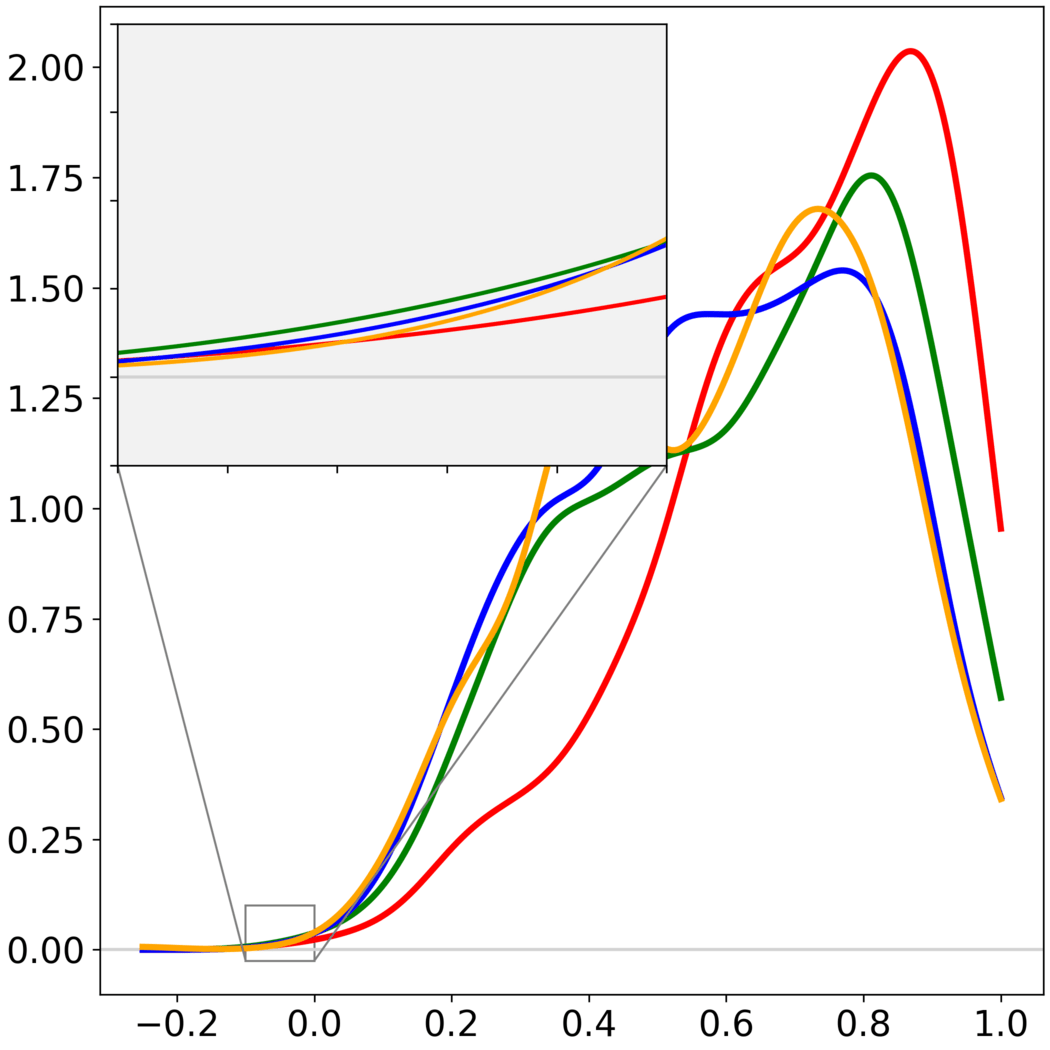}
         \caption{en--de, IT}
         \label{en-de-it-kde-ref}
     \end{subfigure}
     \hfill
     \begin{subfigure}[b]{0.24\textwidth}
         \centering
         \includegraphics[width=\textwidth]{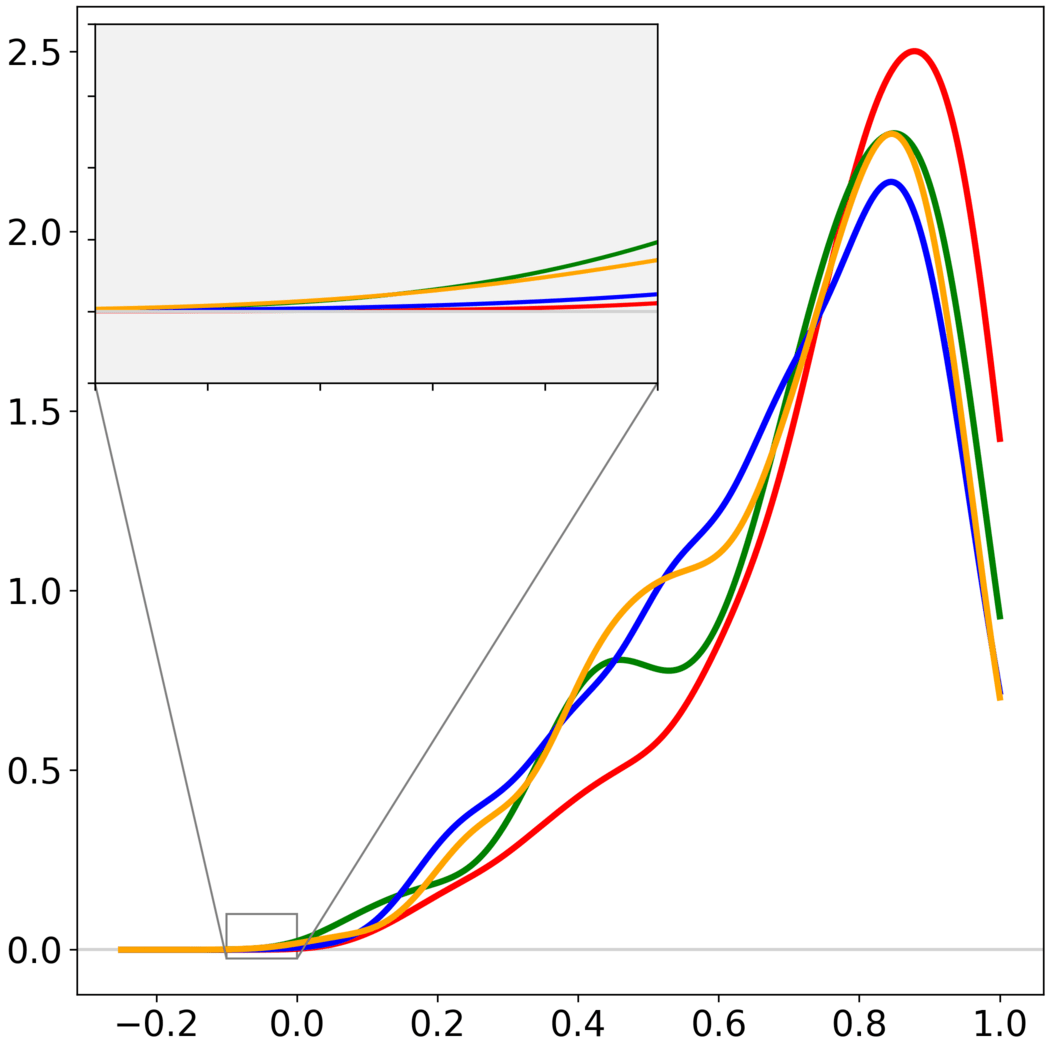}
         \caption{en--de, medical}
         \label{en-de-medical-kde-ref}
     \end{subfigure}
     \hfill
     \begin{subfigure}[b]{0.24\textwidth}
         \centering
         \includegraphics[width=\textwidth]{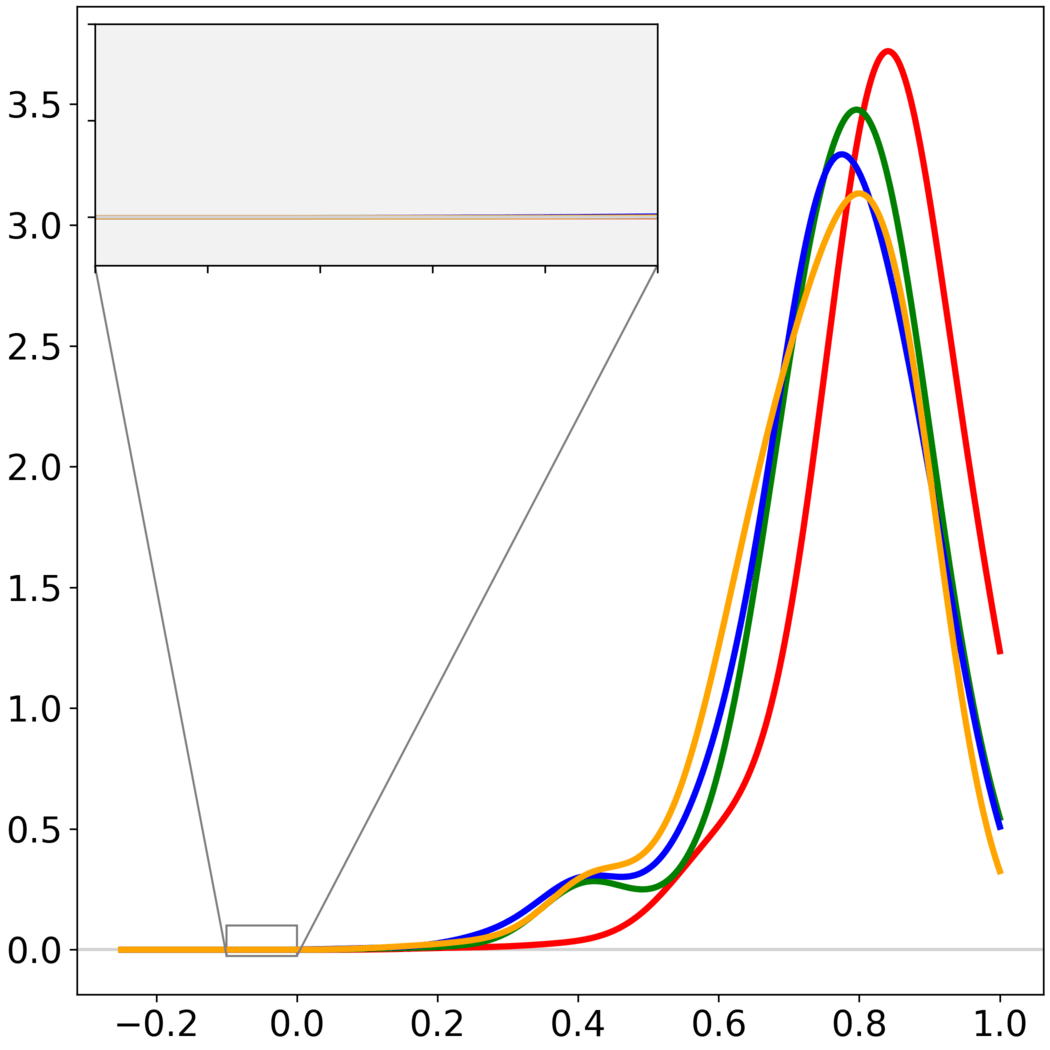}
         \caption{en--de, law}
         \label{en-de-law}
     \end{subfigure}
     \\
     \begin{subfigure}[b]{0.24\textwidth}
         \centering
         \includegraphics[width=\textwidth]{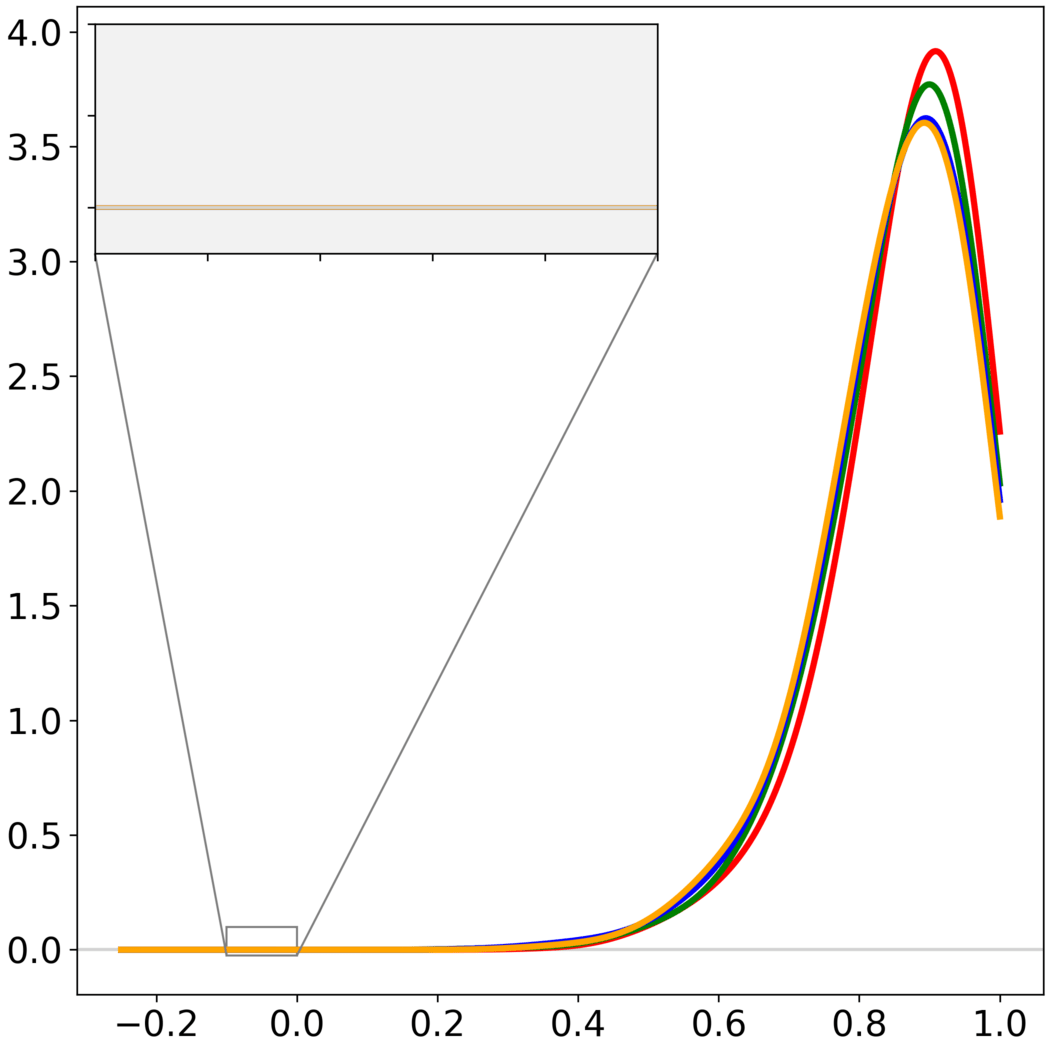}
         \caption{de--en, in-domain}
         \label{de-en-in-domain-kde-ref}
     \end{subfigure}
     \hfill
     \begin{subfigure}[b]{0.24\textwidth}
         \centering
         \includegraphics[width=\textwidth]{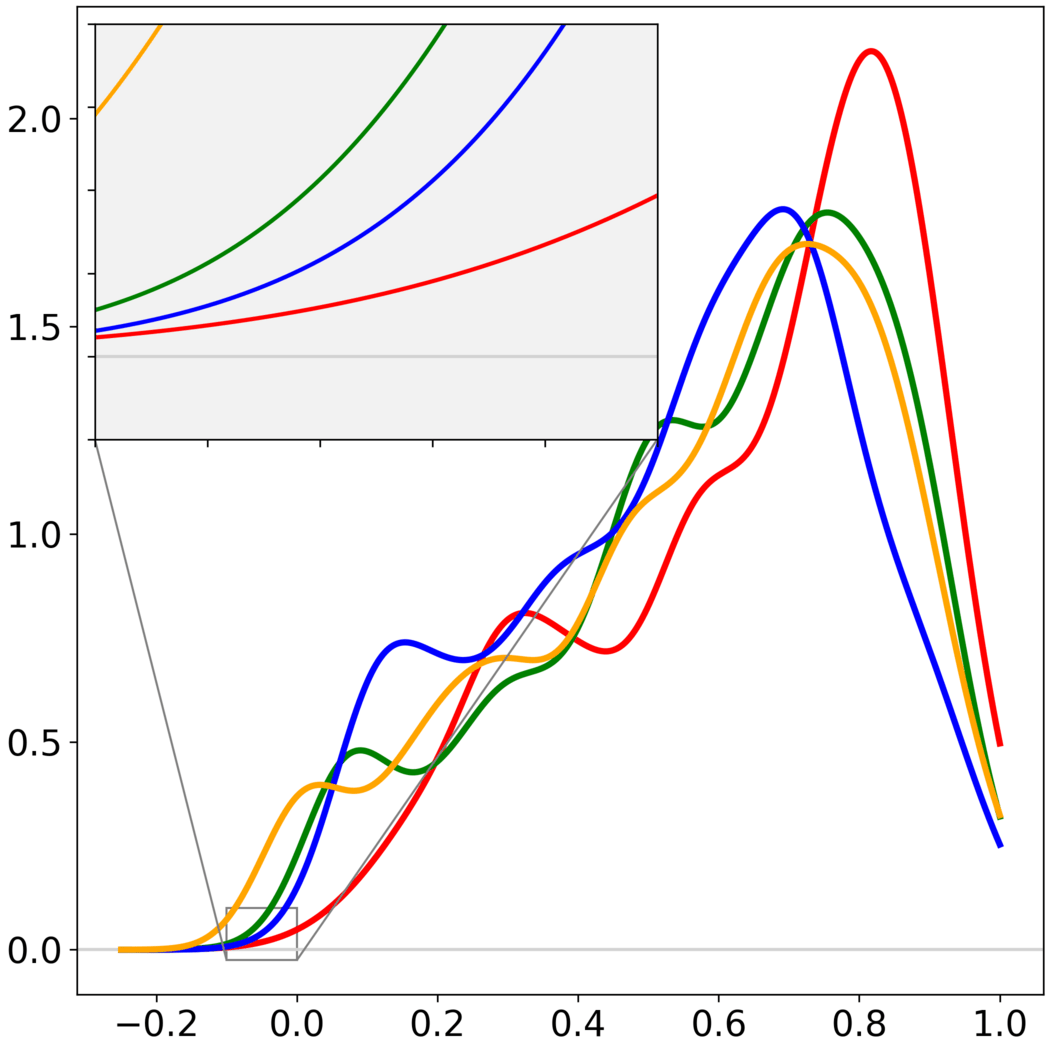}
         \caption{de--en, IT}
         \label{de-en-it-kde-ref}
     \end{subfigure}
     \hfill
     \begin{subfigure}[b]{0.24\textwidth}
         \centering
         \includegraphics[width=\textwidth]{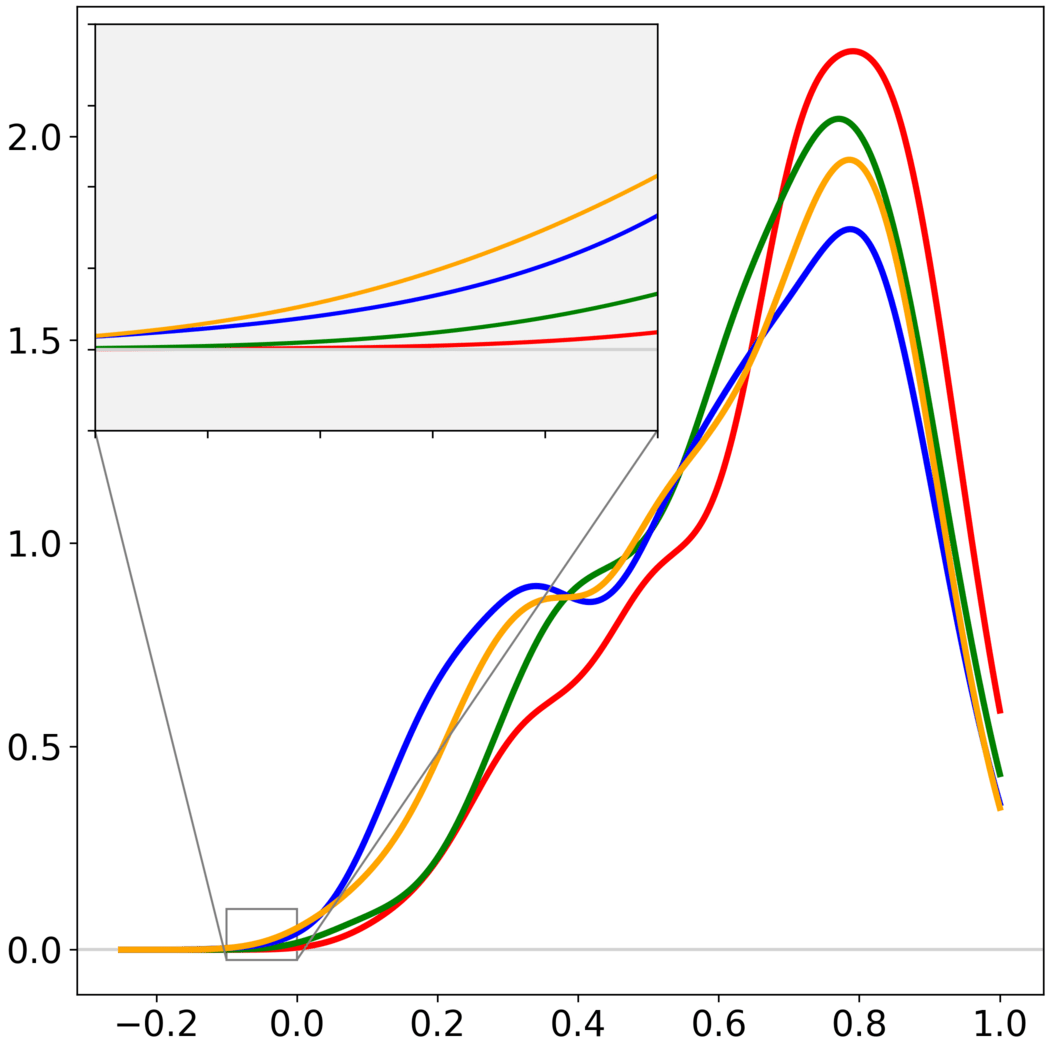}
         \caption{de--en, medical}
         \label{de-en-medical-kde-ref}
     \end{subfigure}
     \hfill
     \begin{subfigure}[b]{0.24\textwidth}
         \centering
         \includegraphics[width=\textwidth]{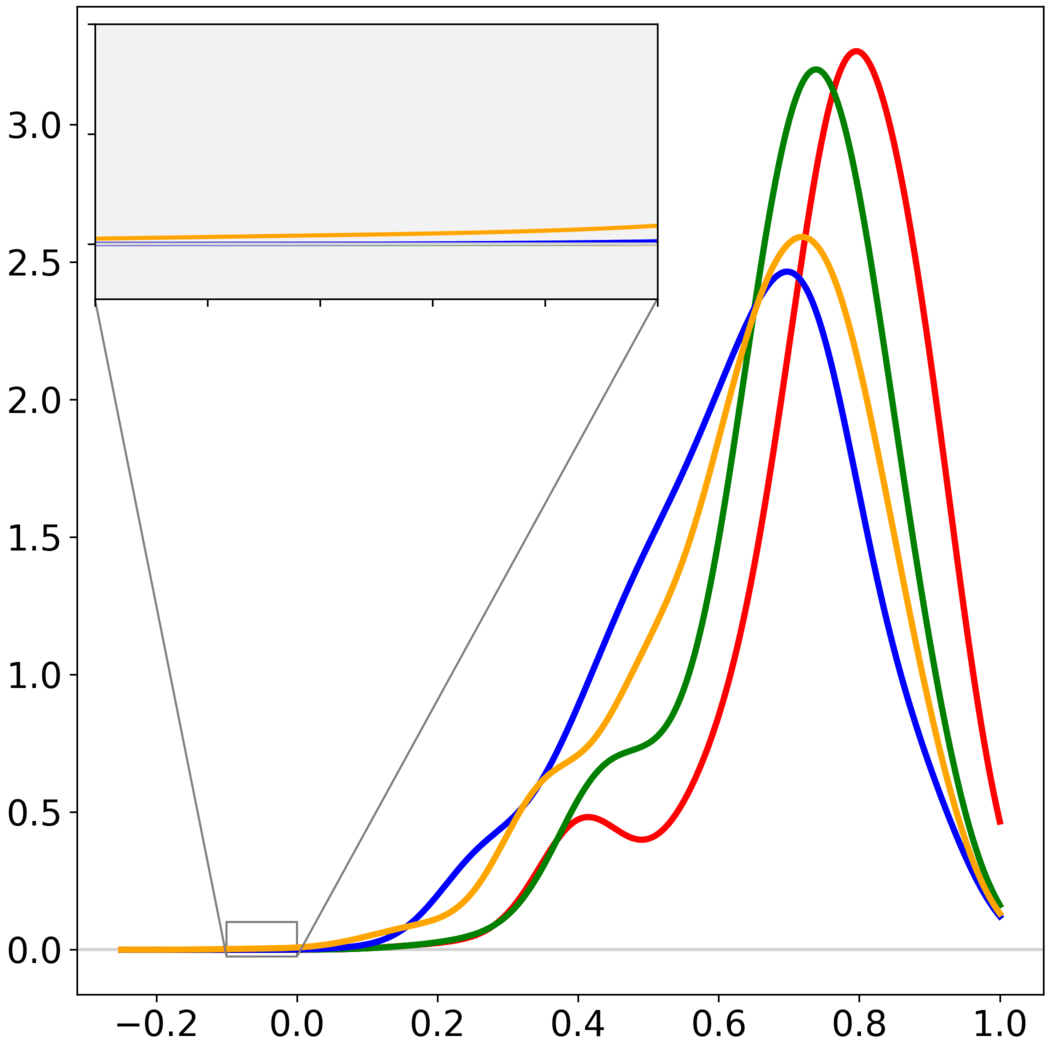}
         \caption{de--en, law}
         \label{de-en-law-kde-ref}
     \end{subfigure}
        \caption{Kernel density estimations (bandwidth=$0.06$) for LaBSE-based cosine similarities between the output produced by NMT models trained in low-resource conditions and the reference translations in test sets belonging to different domains. DA methods: \raisebox{0.05cm}{\textcolor{orange}{\rule{0.5cm}{0.08cm}}}~baseline, \raisebox{0.05cm}{\textcolor{blue}{\rule{0.5cm}{0.08cm}}}~SwitchOut+RAML, \raisebox{0.05cm}{\textcolor{LimeGreen}{\rule{0.5cm}{0.08cm}}}~SeqMix, \raisebox{0.05cm}{\textcolor{red}{\rule{0.5cm}{0.08cm}}}~MaTiLDA.}
        \label{fig:reflabse-low}
\end{figure*}

\begin{figure*}[tb]
\centering
     \begin{subfigure}[b]{0.24\textwidth}
         \centering
         \includegraphics[width=\textwidth]{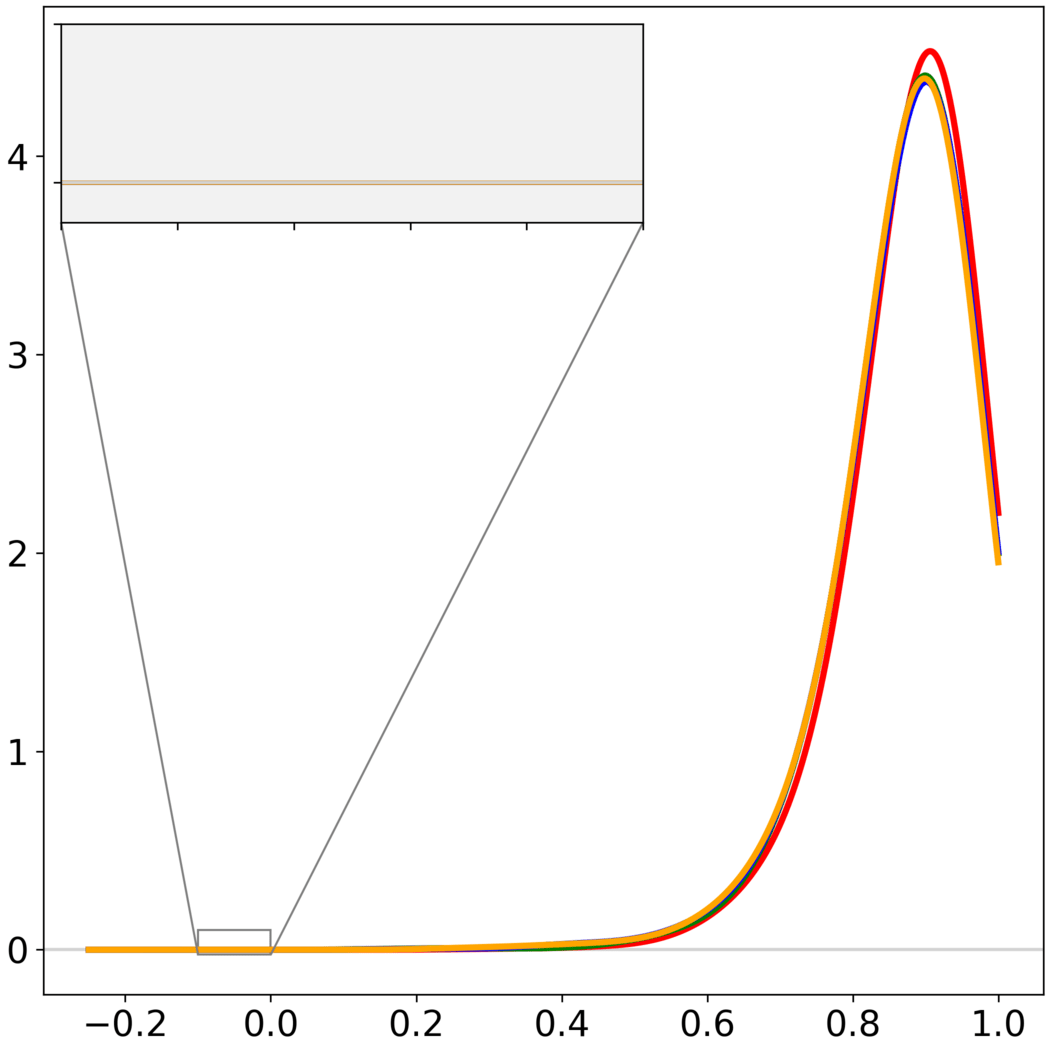}
         \caption{en--de, in-domain}
         \label{large-en-de-in-domain-kde-ref}
     \end{subfigure}
     \hfill
     \begin{subfigure}[b]{0.24\textwidth}
         \centering
         \includegraphics[width=\textwidth]{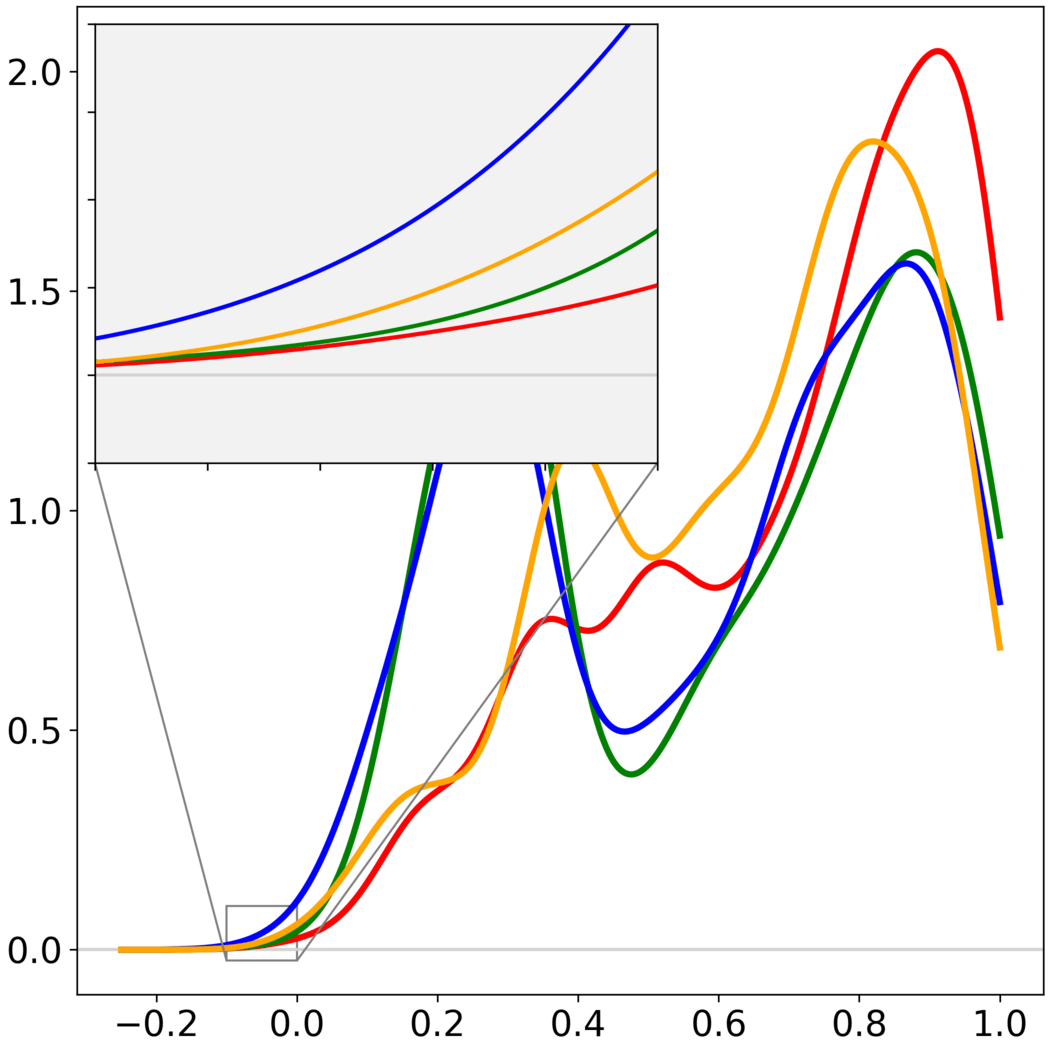}
         \caption{en--de, IT}
         \label{large-en-de-it-kde-ref}
     \end{subfigure}
     \hfill
     \begin{subfigure}[b]{0.24\textwidth}
         \centering
         \includegraphics[width=\textwidth]{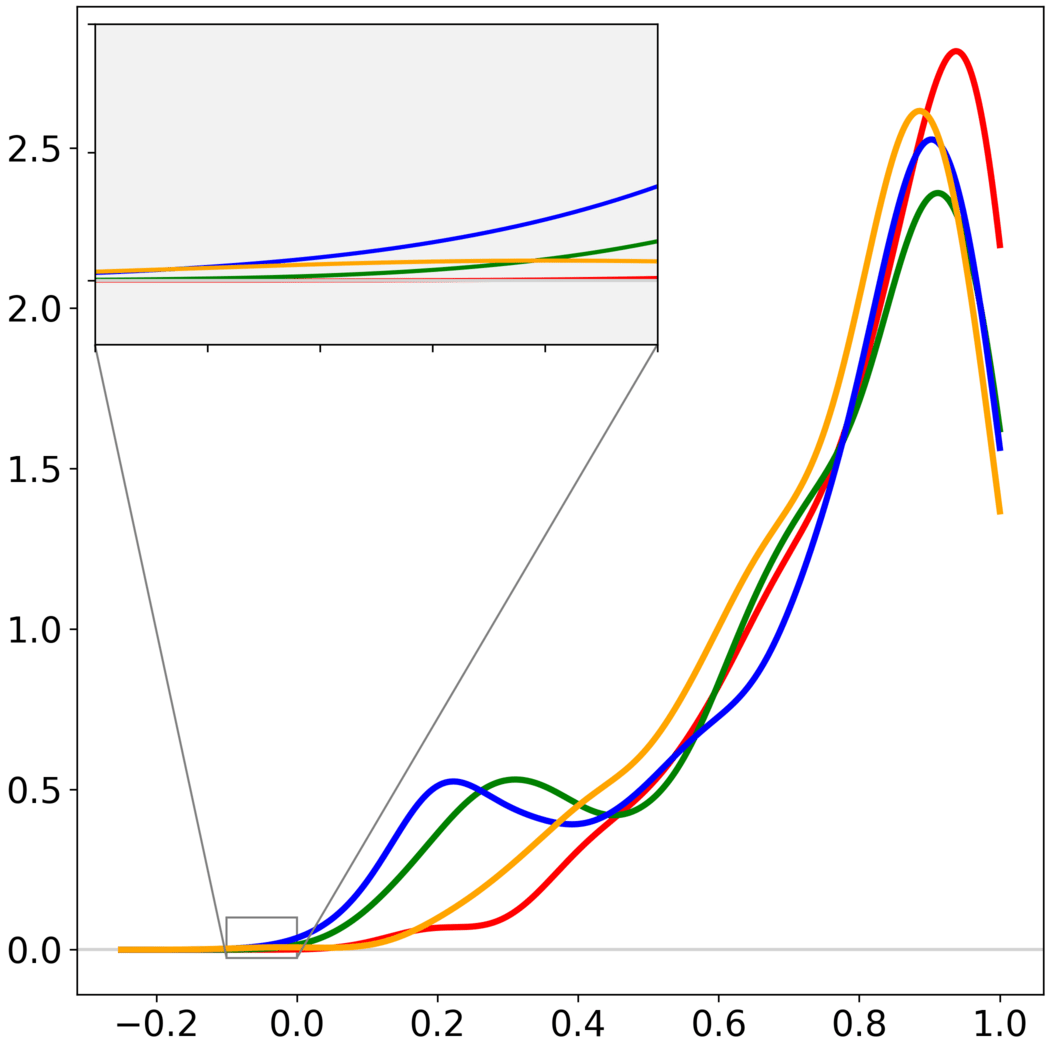}
         \caption{en--de, medical}
         \label{large-en-de-medical-kde-ref}
     \end{subfigure}
     \hfill
     \begin{subfigure}[b]{0.24\textwidth}
         \centering
         \includegraphics[width=\textwidth]{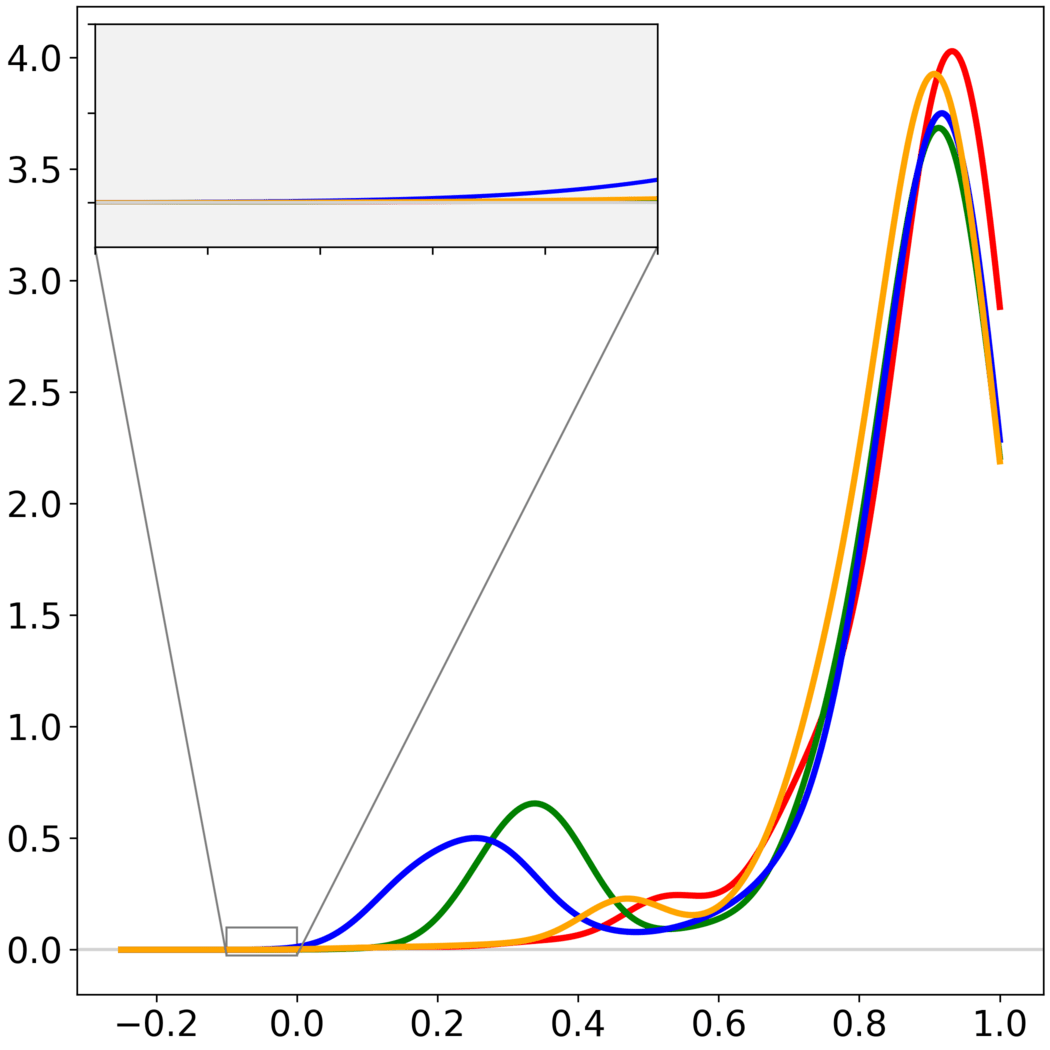}
         \caption{en--de, law}
         \label{large-en-de-law}
     \end{subfigure}
     \\
     \begin{subfigure}[b]{0.24\textwidth}
         \centering
         \includegraphics[width=\textwidth]{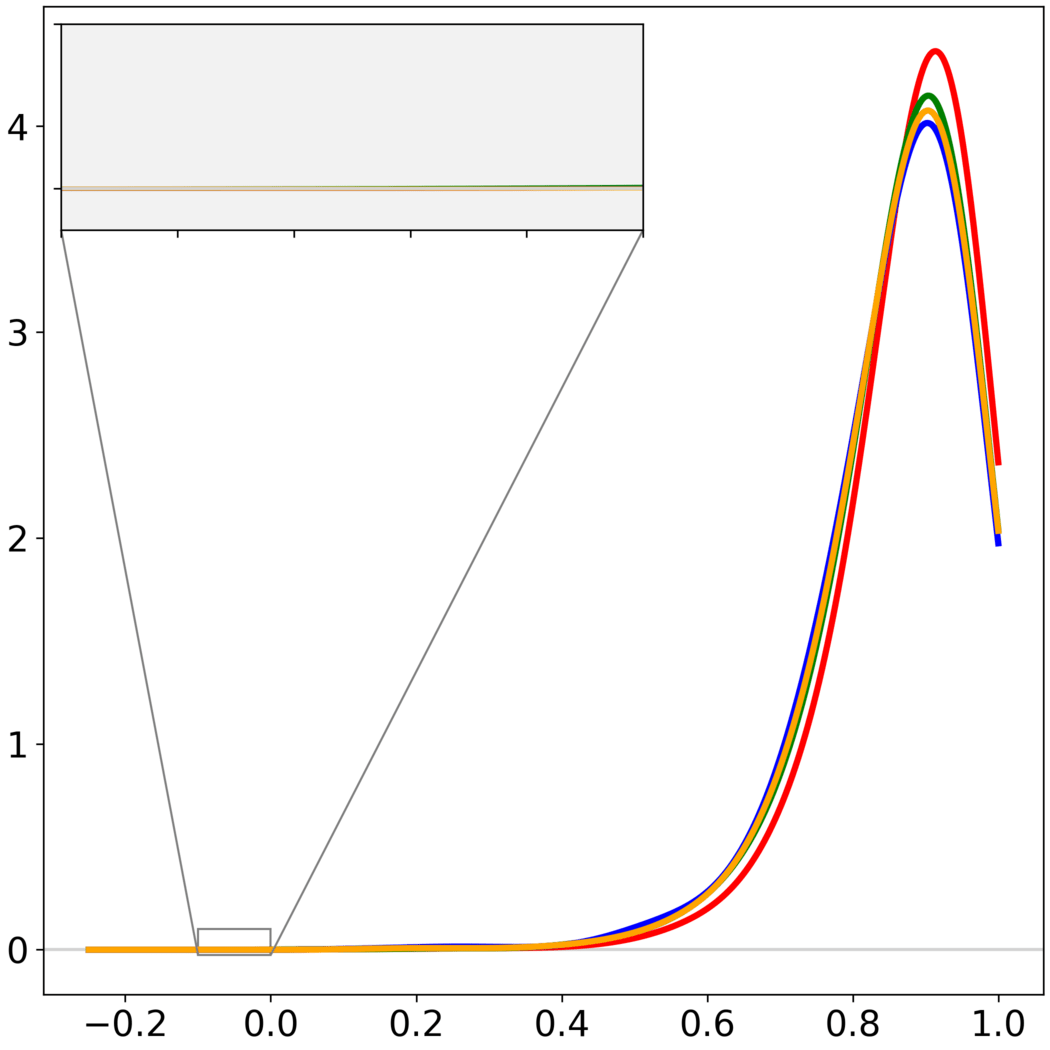}
         \caption{de--en large, in-domain}
         \label{large-de-en-in-domain-kde-ref}
     \end{subfigure}
     \hfill
     \begin{subfigure}[b]{0.24\textwidth}
         \centering
         \includegraphics[width=\textwidth]{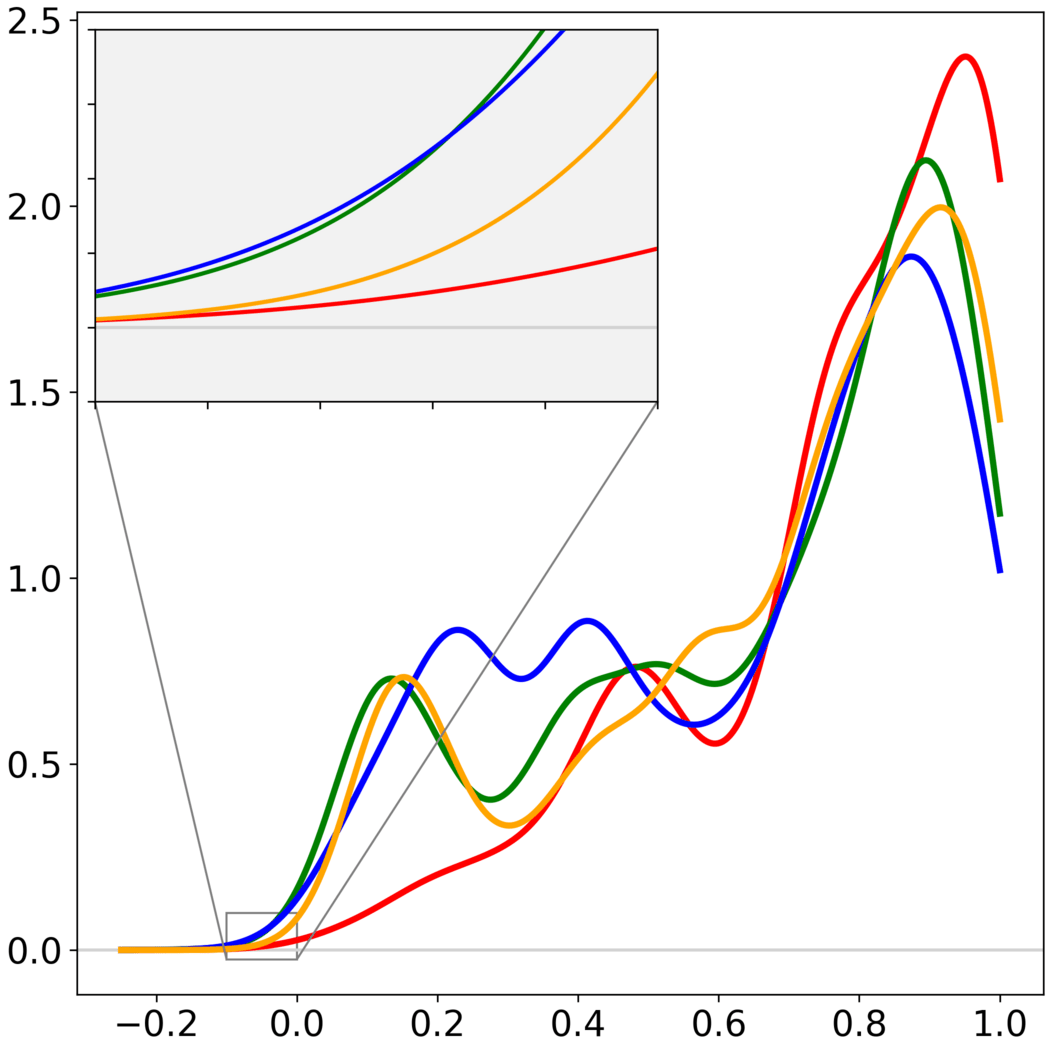}
         \caption{de--en, IT}
         \label{large-de-en-it-kde-ref}
     \end{subfigure}
     \hfill
     \begin{subfigure}[b]{0.24\textwidth}
         \centering
         \includegraphics[width=\textwidth]{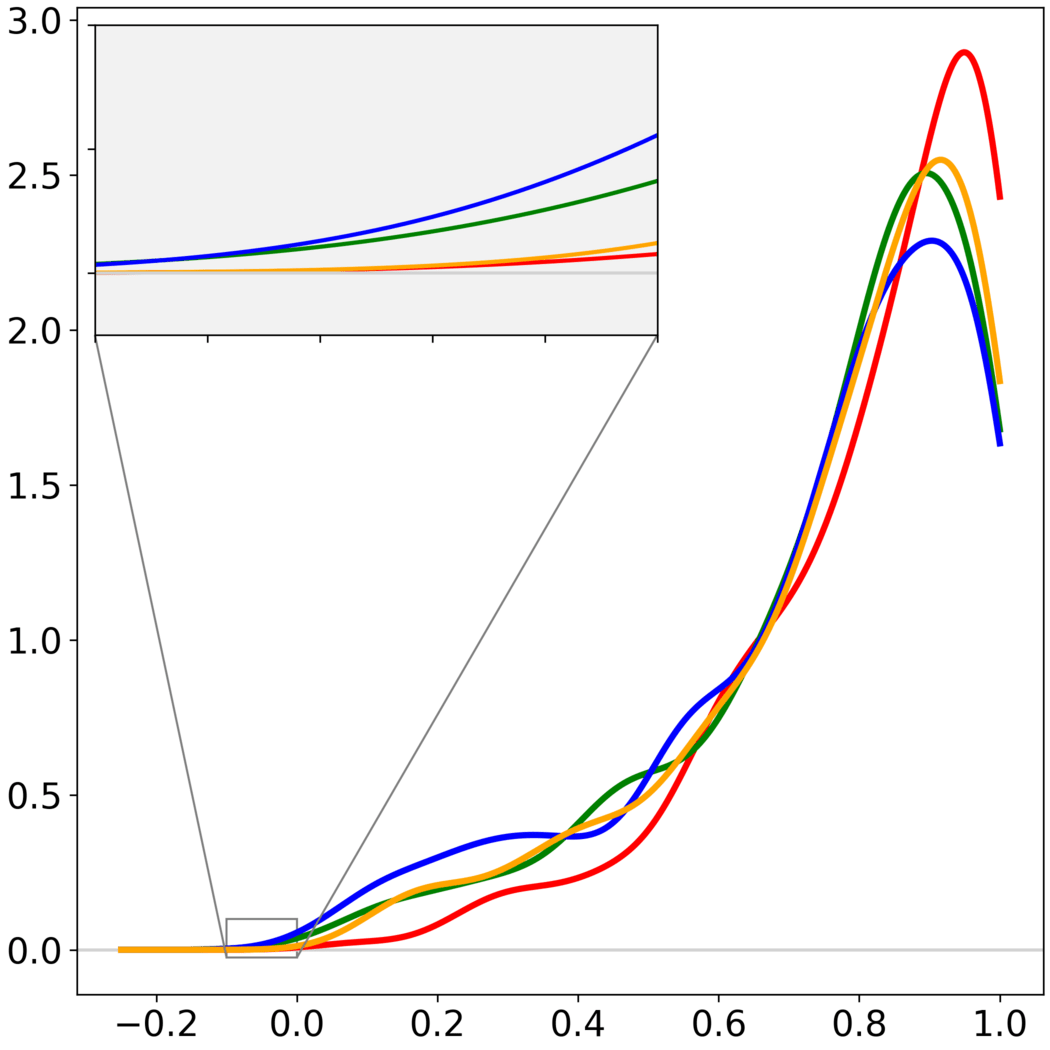}
         \caption{de--en, medical}
         \label{large-de-en-medical-kde-ref}
     \end{subfigure}
     \hfill
     \begin{subfigure}[b]{0.24\textwidth}
         \centering
         \includegraphics[width=\textwidth]{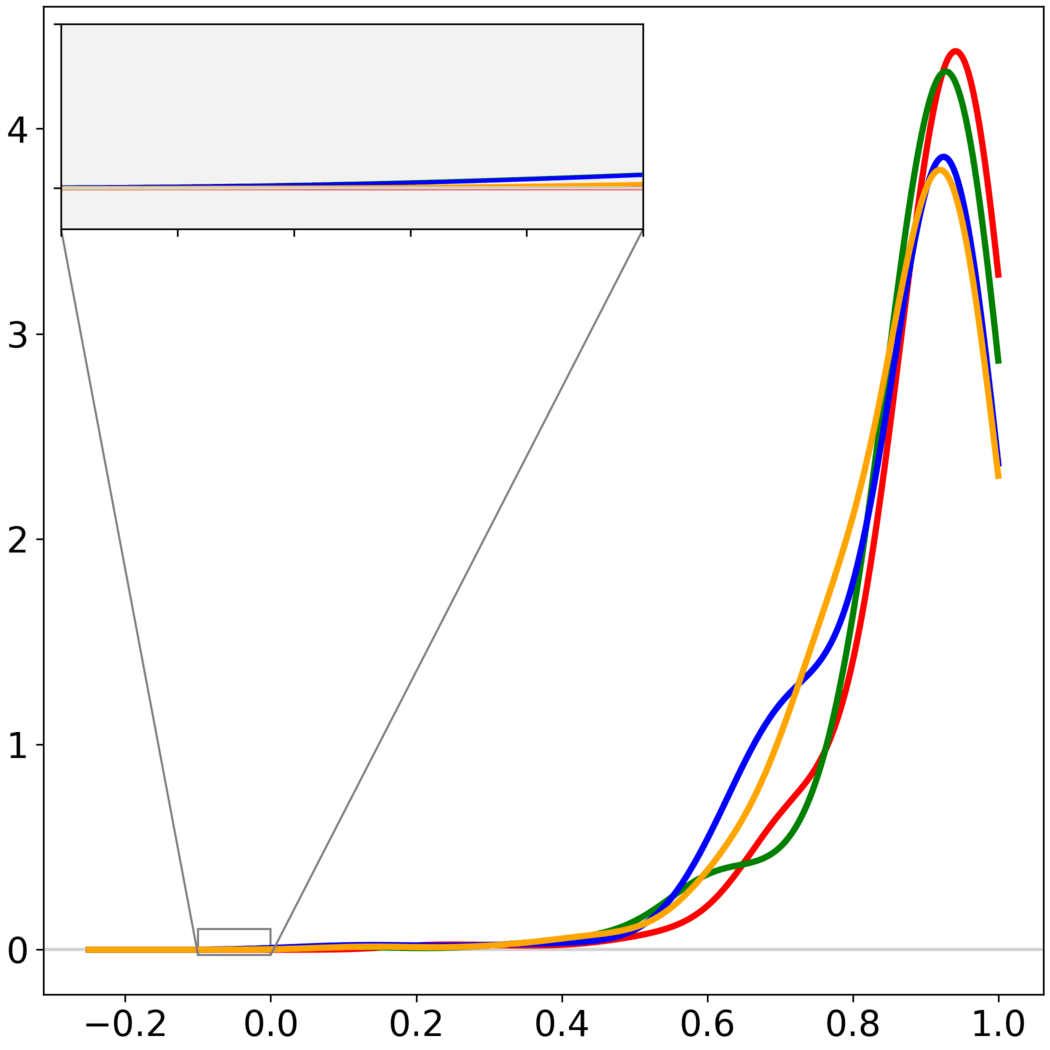}
         \caption{de--en, law}
         \label{large-de-en-law-kde-ref}
     \end{subfigure}
        \caption{Kernel density estimations (bandwidth=$0.06$) for LaBSE-based cosine similarities between the output produced by NMT models trained in high-resource conditions and the reference translations in test sets belonging to different domains. DA methods: \raisebox{0.05cm}{\textcolor{orange}{\rule{0.5cm}{0.08cm}}}~baseline, \raisebox{0.05cm}{\textcolor{blue}{\rule{0.5cm}{0.08cm}}}~SwitchOut+RAML, \raisebox{0.05cm}{\textcolor{LimeGreen}{\rule{0.5cm}{0.08cm}}}~SeqMix, \raisebox{0.05cm}{\textcolor{red}{\rule{0.5cm}{0.08cm}}}~MaTiLDA.}
        \label{fig:reflabse-high}
\end{figure*}

Figures~\ref{fig:reflabse-low} and \ref{fig:reflabse-high} 
show the kernel density estimations of the distributions of cosine similarities for the systems trained in low-resource and high-resource conditions, respectively, for the English--German and German--English translation tasks. To obtain these plots we used the same test sets and domains (in-domain, IT, legal and medical) used 
in Sec.~\ref{se:domain-robust}. 
A larger view of the area close to zero where the strongest hallucinations are supposed to live~\cite{detecting2022} is also included with each plot.

As the plots show, MaTiLDA cosine distribution curves are shifted to the right when compared to the baseline and the other DA methods; even when the systems are trained in high-resource conditions (see Fig.~\ref{fig:reflabse-high}). This can be interpreted as a clear sign of reduction of hallucinations in the systems trained with MaTiLDA, especially under domain shift. In this regard, it is worth noting how the in-domain plots are prominently shifted to the right when compared to the out-of-domain ones because the systems hallucinate less and produce more semantically accurate outputs when translating in-domain texts. Table~\ref{tab:example} shows the output translations of each system under study in a couple of representative cases in which MaTiLDA attains the highest cosine similarity with the reference translation. Note, however, that a relatively high cosine value may still correspond to a hallucination (see cosine similarity values for the second example in Table~\ref{tab:example}), which supports the idea that the consistent shift to the right of the cosine distribution curves for MaTiLDA in figures~\ref{fig:reflabse-low} and \ref{fig:reflabse-high} clearly indicate a reduction in the tendency to hallucinate of the systems trained with MaTiLDA.

\begin{table}[tb]
\centering
\begin{tabular}{llp{5cm}}
\textbf{Type} & \textbf{Cosine} & \textbf{Sentence} \\ \hline\hline
Source & & hinweise fÜr die richtige anwendung \\
Reference & & advice on correct administration \\
Baseline & 0.167 & you know, evidence for die's application. \\
SwOut+RAML & 0.098 & it's called "die." \\
SeqMix & 0.280 & clues to die's real application. \\
MaTiLDA & 0.515 & clues to the right use. \\
 \hline
Source & & artikel 16 \\
Reference & & article 16 \\
Baseline & -0.049 & they said, "wwhwhwhwhwhwhwhwhwhwhwhwhw [...] were were were were were were were were were [...] were." \\
SwOut+RAML & 0.529 & I was 16 years old. \\
SeqMix & 0.634 & it's 16 articles of 16. \\
MaTiLDA & 0.893 & articles 16. \\
 \hline
\end{tabular}
\caption{Examples of output translations in which MaTiLDA attains the highest cosine similarity with the reference translation.} 
\label{tab:example}
\end{table}

\section{Related work}
\label{sec:related}

The back-translation~\cite{sennrich-back} approach for leveraging additional target monolingual data, 
is, probably, the most popular DA approach for NMT. The set of related approaches covered in this section, however, mainly focus on methods that, as MaTiLDA, do not require additional resources besides the training parallel corpus. 

Li et~al.~\cite{li-etal-2019-understanding-data} evaluate back- and forward-translation in such a setting. They train forward and backward NMT systems on the available parallel data and use them to produce new synthetic samples by translating either the target side~\cite{sennrich2016edinburgh} or the source side~\cite{zhang-zong-2016-exploiting} of the original training corpus. Other approaches simply select two training samples at random and concatenate, on one side, the source sentences and, on the other side, the target sentences to generate larger training samples \cite{kondo-etal-2021-sentence,wu-etal-2021-mixseq} in order to improve the translation quality of long source sentences.

The approaches we have evaluated in our experiments, RAML~\cite{norouzi2016reward}, SwitchOut~\cite{wang-etal-2018-switchout} and SeqMix~\cite{guo-etal-2020-sequence}, aim at extending the support of the empirical data distribution, which is expected to prevent the model to memorize long segments and improve the model generalization capabilities.
To that end, RAML and SwitchOut replace words with other words sampled from a uniform distribution over the vocabulary, which, in practice, results in infrequent words being overrepresented; RAML works on the target side, whereas SwitchOut works on the source side. SeqMix approaches the problem in a different way and generates synthetic training samples by randomly combining parts of two sentences in order to encourage compositional behaviour.


Applying simple transformations, such as swapping words, to existing sentences is an idea that has been widely applied in the context of NMT with multiple purposes. For instance, several simple transformations (word deletion, replacement, swapping) have been applied to back-translated data~\cite{edunov-etal-2018-understanding} with the aim of allowing the system to better distinguish between original 
and back-translated sentences.
In the context of unsupervised NMT, the noisy input sentences to denoising autoencoders are generated by random token swaps~\cite{artetxe2017unsupervised}. 

Focusing on approaches that modify word order in the context of DA,
it is worth highlighting 
a self-translation approach using a right-to-left decoder~\cite{zhang2019regularizing}, which is similar to inverting the order of the target words. However, unlike MaTiLDA, this last approach needs to generate translations from the model during training. 

Replacing tokens with placeholders (as we do in \emph{unk}) has already been applied to the source language~\cite{zhang2020token} in combination with two self-supervised learning objectives for detecting replaced and dropped tokens. Xie et al.~\cite{xie2017data} also evaluate the impact of random replacements of words in the source and target sides of the training samples by either a random word from the vocabulary, or by a blank.

Gao et~al.~\cite{gao2019soft} replace \miqueladd{source-side} words selected at random with \emph{soft words} whose representations are obtained from the probability distribution provided by a language model. 
Fadaee et~al.~\cite{Fadaee_2017} replace some words in their training samples by infrequent words in order to improve the performance of the NMT model when dealing with them at translation time. 
Words to be replaced are identified using
a large source language model. 
Once the source words to be replaced are identified, a word-alignment model and a probabilistic dictionary are used to also replace the corresponding counterpart by the most probable translation of the new source word. In MaTiLDA, the \emph{replace} transformation, which is similar, 
does not require any language model.  


As regards the special token we use to prevent negative transfer between tasks, a similar strategy \cite{caswell2019tagged} has been applied to identify synthetic samples when combining actual parallel data and back-translated data for training. Yang et~al.~\cite{YANG2019240} extends this  last work by including forward-translated data for training using two different special tokens to distinguish the two types of synthetic data. 
Another strategy that has been reported to be effective to combine synthetic and original training instances is the AugMix method~\cite{hendrycks2020augmix}, initially defined in the context of image processing. This method involves creating training samples through linear interpolation of the embeddings of both the original and the synthetic samples and adding an auxiliary loss that enforces model probabilities of both types of samples to be similar. It has been subsequentially applied to NMT~\cite{jin2022admix} using
the simple transformations \miqueladd{mentioned above} (word deletion, replacements as those defined in SwitchOut/RAML, swapping), which are not the best performing ones according to our analyses.

The problem of the NMT system relying too much on the target-language context has been addressed in ways other than DA.
Miao et~al.~\cite{miao-etal-2021-prevent} define a metric to measure the prevalence of the decoder's language model over the 
encoder representations 
and use it to define specific auxiliary loss functions to reduce this prevalence;
Weng et~al.~\cite{weng2020towards} use a similar auxiliary loss that is only optimized on mistranslated fragments selected from the training data.

Finally, a number of approaches that mitigate the amount of hallucinations produced by NMT systems have been proposed~\cite{needle2022,hallucinations2018,muller-etal-2020-domain,muller-sennrich-2021-understanding,wang-sennrich-2020-exposure}. 
However most of them either do not evaluate the impact of the  techniques proposed in the general performance of the NMT models built~\cite{needle2022,hallucinations2018}, or report mixed results with a drop in performance for some language pairs or in some translation scenarios~\cite{muller-etal-2020-domain,muller-sennrich-2021-understanding,wang-sennrich-2020-exposure}.
In contrast, the approach described in this paper not only reduces the degree of hallucinations as measured by LaBSE, but it also improves the general quality of the translations produced by the NMT models.

\section{Concluding remarks}\label{sec:conclusions}
We have presented a novel method for data augmentation (DA) for neural machine translation (NMT) that we have termed as multi-task learning DA (MaTiLDA). In contrast to state-of-the-art DA approaches, MaTiLDA aims at generating new synthetic training samples with non-fluent target-language sentences by means of aggressive transformations, such as reversing the order of the target sentence or swapping random target words. The new synthetic training samples, which are considered as data for additional learning tasks, provide new contexts during training where the target prefix is not sufficiently informative to predict the next token, thus strengthening the relevance of the encoder and increasing at inference time the reliance on the source-language representations it generates. MaTiLDA is agnostic to the NMT model architecture and does not require elaborate preprocessing steps, additional training systems, or data besides the available training parallel corpora.

We have extensively evaluated this new approach on ten low-resource and four high-resource translation tasks. The results show consistent improvements over a baseline without DA, and over three strong state-of-the-art DA methods that aim at extending the support of the empirical data distribution by generating synthetic training samples with fluent target sentences. This improvement shows up both when training NMT systems in low- and high-resource conditions. 
Furthermore, NMT systems trained with MaTiLDA are much more robust under domain shift and generate fewer hallucinations than the baseline or any of the state-of-the-art DA methods we have compared with when translating out-of-domain texts. In addition, our evaluation also demonstrates that MaTiLDA can be easily combined with the standard DA method, namely back-translation, and that both methods complement each other as their combination results in further translation performance improvements. 

An analysis of the influence of the encoder and decoder representations in the NMT system output shows that, thanks to the transformations used for building synthetic training samples, MaTiLDA increases the contribution of the source representations from the encoder to the decisions made by the NMT decoder during inference. Hence, MaTiLDA makes it possible to build NMT models for low-resource language pairs that, even though they have been trained on small parallel corpora, are able to behave as if they had been trained on larger training corpora.

All in all, the method we have presented offers promising implications for the field of NMT. By utilizing a robust and contrasted approach, we have demonstrated the potential to enhance the accuracy of virtually any existing NMT system by seamlessly integrating MaTiLDA in their training pipelines to make the most of the existing corpora. \revadd{We leave for future work the study of the potential effects of integrating MaTilDA in scenarios 
with even more resources than those used in our study, as it is the case of the utilization of large pre-trained multilingual NMT models. This pre-trained models make use of extensive amounts of monolingual and parallel data from various language pairs and significantly outperform systems trained exclusively on parallel data~\cite{tang-etal-2021-multilingual}}.\footnote{Indeed, the performance gap between our baseline systems and the version with 1.3B parameters of the pre-trained model NLLB \cite{nllbteam2022language} when it is fine-tuned on the parallel corpora used in our low-resource experiments is, on average, 7.4 BLEU points when translating into the low-resource languages, and 10.4 BLEU points when translating into the high-resource languages (English or German).}

\section*{Acknowledgments}
This paper is part of the R+D+i project PID2021-127999NB-I00 funded by the Spanish Ministry of Science and Innovation (MCIN), the Spanish Research Agency (AEI/10.13039/501100011033) and the European Regional Development Fund A way to make Europe. The computational resources used were funded by the European Regional Development Fund through project IDIFEDER/2020/003. We thank Mikel L. Forcada for his comments on a draft version of this paper.


\ifCLASSOPTIONcaptionsoff
  \newpage
\fi



\bibliographystyle{IEEEtran}
\bibliography{IEEEabrv,emnlp}

\begin{IEEEbiography}[{\includegraphics[width=1in,height=1.25in,clip,keepaspectratio]{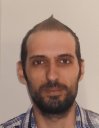}}]{Víctor M. Sánchez-Cartagena} is Assistant Professor at Universitat d’Alacant, Spain. He
obtained his PhD in Computer Science in 2015, he also worked as a research engineer at Prompsit Language Engineering.
His main fields of research are deep learning and machine translation with an emphasis on low-resource languages and the combination of multiple systems and/or sources of information. He authored more than 10 indexed publications, including top natural language processing conferences.
\end{IEEEbiography}

\begin{IEEEbiography}[{\includegraphics[width=1in,height=1.25in,clip,keepaspectratio]{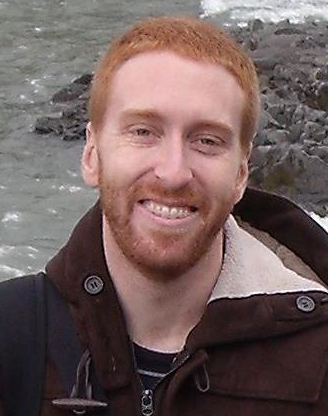}}]{Miquel Esplà-Gomis} is Assistant Professor at Universitat d’Alacant, Spain. He
obtained his PhD in Computer Science in 2016. His main research fields are parallel data acquisition and application of translation technologies to computer-aided translation. He has published more than 30 articles in 
international conferences and journals. He has coordinated the EU-funded project MaCoCu, aimed at harvesting monolingual/parallel corpora for low-resourced European languages from the Internet.
\end{IEEEbiography}

\begin{IEEEbiography}[{\includegraphics[width=1in,height=1.25in,clip,keepaspectratio]{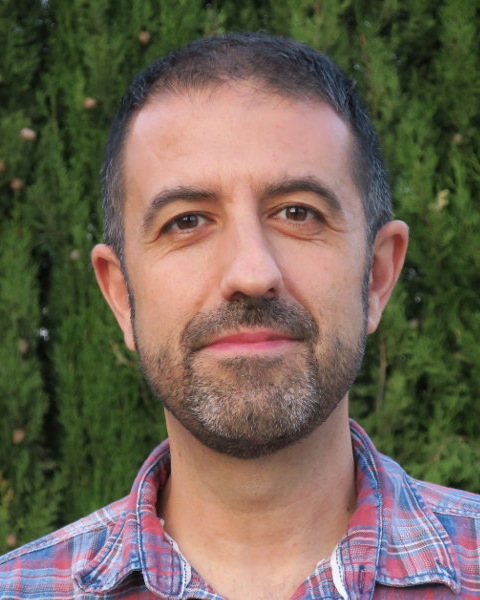}}]{Juan Antonio Pérez-Ortiz} is Associate Professor of computer science at Universitat d’Alacant, Spain, director of the Transducens research group, and co-founder of Prompsit Language Engineering. He has worked on machine translation (rule-based, statistical and neural) and computer-aided translation since 1999, especially as a member of the team involved in the development of the Apertium platform. 
His current research focuses on neural language technologies for low-resource languages.
\end{IEEEbiography}

\begin{IEEEbiography}[{\includegraphics[width=1in,height=1.25in,clip,keepaspectratio]{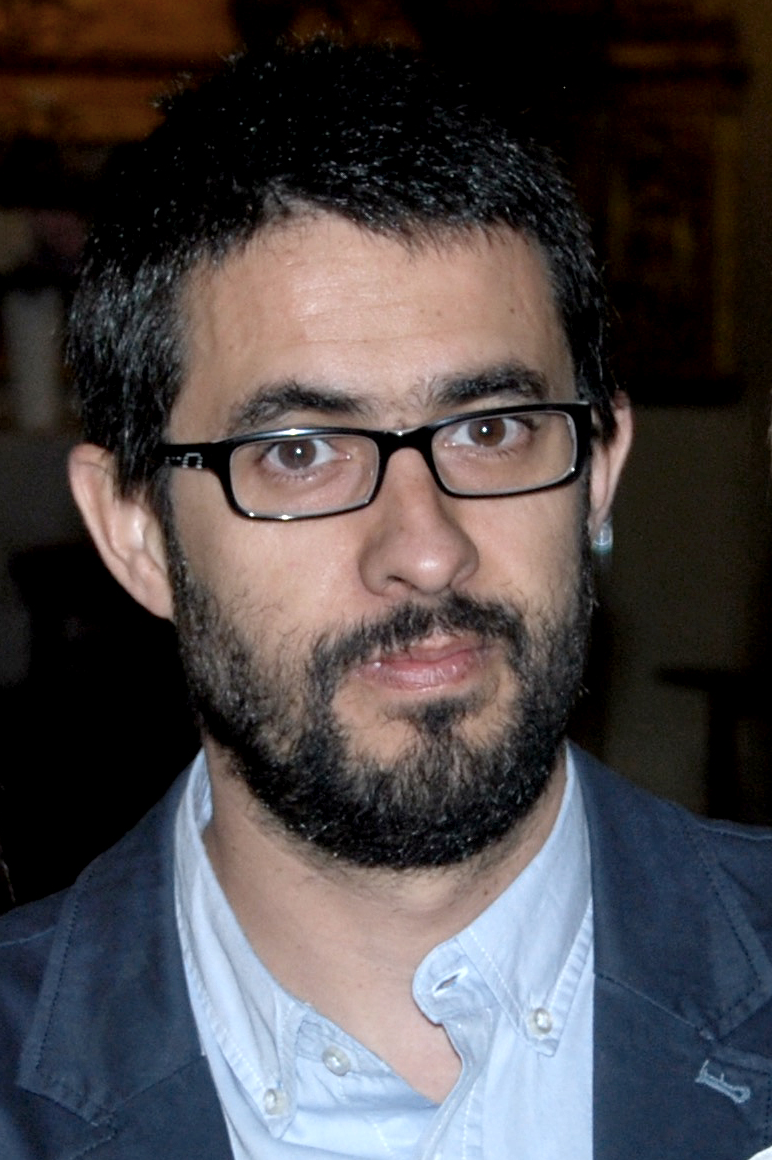}}]{Felipe Sánchez-Martínez} is Associate Professor at Universitat d'Alacant, Spain. His main field of research is on low-resource machine translation and the integration of machine translation in other translation technologies. He has participated in several research projects funded by the Spanish Government and the European Commission.  He  contributed to the design and  development of the Apertium shallow-transfer machine translation platform, and co-founded the company Prompsit Language Engineering.
\end{IEEEbiography}

\end{document}